\documentclass{article} 
\usepackage{iclr2027_conference,times}

\usepackage{amsmath,amsfonts,bm}

\def\eqref#1{equation~\ref{#1}}

\def\1{\bm{1}}

\DeclareMathAlphabet{\mathsfit}{\encodingdefault}{\sfdefault}{m}{sl}
\SetMathAlphabet{\mathsfit}{bold}{\encodingdefault}{\sfdefault}{bx}{n}

\usepackage{float}
\usepackage{wrapfig}
\usepackage{needspace}
\usepackage{hyperref}
\usepackage{url}
\usepackage{graphicx}
\usepackage{booktabs}
\usepackage{multirow}
\usepackage{xcolor}
\usepackage{colortbl}
\definecolor{HeaderColor}{RGB}{226,230,236}
\definecolor{PrintedTint}{RGB}{248,242,228}
\definecolor{MaskedTint}{RGB}{234,241,238}
\definecolor{AltTint}{RGB}{232,242,247}
\definecolor{FailTint}{RGB}{246,232,225}
\definecolor{PassTint}{RGB}{230,239,229}
\usepackage{amsmath}
\usepackage{amssymb}
\usepackage{pifont}

\title{CARAT: Do Materials LLMs Reason or Recite?}

\author{Jiajun Wu\textsuperscript{1}, Jian Yang\textsuperscript{1}, Zixiang Ni\textsuperscript{2}, Zhenzhu Li\textsuperscript{3}, Bin Chong\textsuperscript{4}\\
\textsuperscript{1}Beihang University\\ \textsuperscript{2}Xi'an Jiaotong University\\ \textsuperscript{3}Imperial College London\\ \textsuperscript{4}Peking University}

\iclrfinalcopy
\begin{document}
\raggedbottom

\maketitle

\begin{abstract}
When a materials LLM answers a question about crystal structure, does it
reason from the structure or copy an answer already printed in its
input? Accuracy cannot tell: a structural description often prints the very field it is scored against. \textbf{CARAT} holds question and gold
answer fixed across eight matched views, names each structural relation
separately in GraphSpace, and adds matched fine-tuning, answer masking,
evidence injection, paired inference, and a rule that can withhold claims. First, on the benchmark's hardest
families the grounded view is
worth 17.3 points over formula inputs. Second, we turn that scrutiny on
ourselves. GraphSpace beats a plain periodic graph by 19.3 points, but
that margin is two effects at once: where the plain rendering carries
everything the question needs it is 1.96 points, and where it omits
those fields entirely, 46.7 points. The headline mostly measures what
the baseline lacked, not how evidence is presented. Third, we attack our
own benchmark. A rule that skips the link and reads the list directly
answers four of seven hardened families, so we rebuilt it until eleven
such shortcuts sat near chance. The frozen model quotes that link yet
answers the same when we redirect it, on 95.6\% of paired cases: it
repeats the relation without using it. After matched supervision it
reaches 99.8\%, and deleting the link drops it to 23.4\%, below the
27.0\% the best shortcut reaches: both steps are learnable.
\end{abstract}

\begin{figure}[H]
\centering
\vspace{-10pt}
\includegraphics[width=0.8\textwidth,trim=0 0 0 9.8bp,clip]{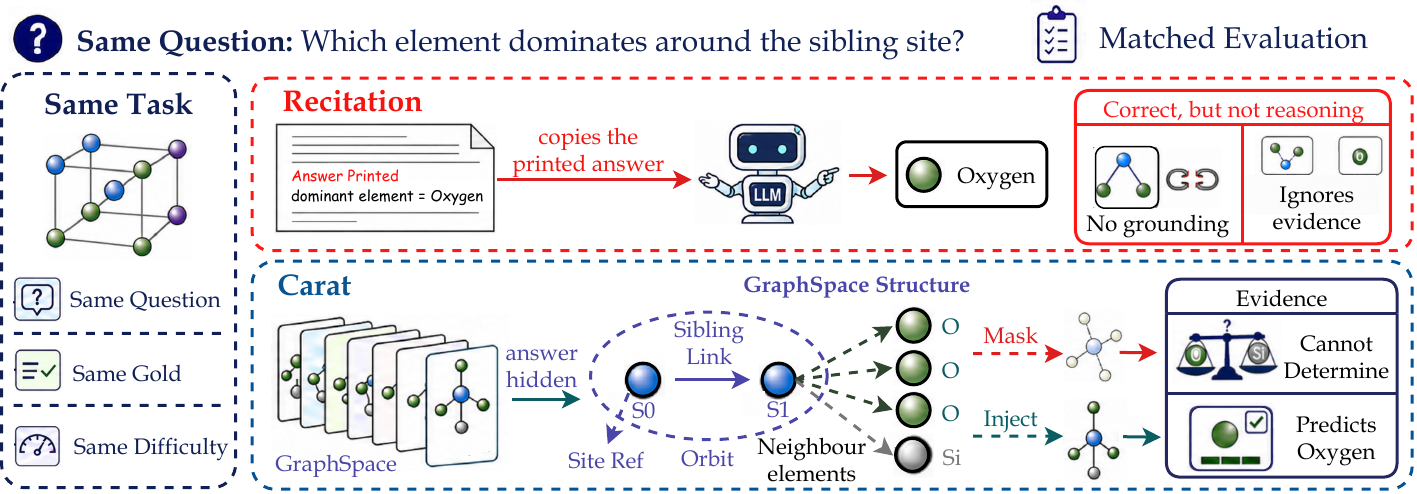}
\vspace{-10pt}
\begin{minipage}{0.8\textwidth}
\caption{\textbf{Reasoning or recitation?} CARAT varies structural evidence under matched questions and tests answer dependence through masking and injection.}
\label{fig:carat-intro}
\end{minipage}
\end{figure}

\section{Introduction}
\label{sec:introduction}

Graph neural networks predict crystalline materials properties directly from
periodic crystal graphs
\citep{xie2018cgcnn,chen2019megnet,choudhary2021alignn}. Large language models
offer a different interface: they consume textual descriptions of structure
and answer questions about properties and local environments
\citep{rubungo2025llmprop,gruver2024crystalllm,jablonka2024gpt}. Near-perfect scores on symmetry, coordination or dimensionality questions
therefore look like structural reasoning. But do those answers actually
depend on the structural relations, or only on how the relations happen to
be printed?

The distinction is hard to make because a structural description often
already contains the answer that the task asks for. Robocrystallographer
\citep{ganose2019robocrys} prints the space group in plain text; a model that
returns that group may infer it from atomic positions or simply copy the
supplied string, and both behaviours receive the same score. We call this \emph{readback}: the input prints the gold answer, so a correct
response need not involve any inference. It is a materials-specific instance
of shortcut learning \citep{geirhos2020shortcut}, and standard evaluation
cannot resolve it. A single fixed format cannot separate evidence use from
answer extraction. Deleting an answer-bearing field still leaves
tolerance-dependent labels and surface shortcuts. An aggregate gain need not
survive pairing over the same materials. The remedy is to change the
evidence while the question, gold answer and difficulty stay fixed, then
test which changes alter the answer. Figure~\ref{fig:carat-intro}
shows this for one such question, about a neighbour reached through a
sibling site.

We implement this design in CARAT, a \textbf{C}ontrolled \textbf{A}blation of
structural \textbf{R}epresentations with \textbf{A}nti-shortcut
\textbf{T}esting. Each question is rendered into eight matched views: a
chemical formula, raw CIF, lattice and sites, a plain periodic graph,
GraphSpace, and three GraphSpace ablations that remove graph, symmetry, or
citation information. Shared scaffolds keep unrelated wording fixed, paired evaluation tracks the effect of each evidence change on the
same underlying task, an anti-shortcut battery tests answer readback,
tolerance chasing, and surface-pattern matching, and a paired statistical
layer adds material-cluster bootstrap intervals, multiple-comparison
correction, and a preregistered decision rule. Answer-printing cases are kept as readback diagnostics but excluded from the
hardened capability metric, so an extracted label is never counted as
structural inference. Figure~\ref{fig:carat-framework} ties these controls to
representation-matched training and mask-and-inject tests.

On a benchmark of 1,024 matched questions over 562 materials, a 27B model
gives GraphSpace a hard-slice aggregate advantage of 17.3 percentage points
over formula inputs, 95\% CI $[15.8,18.7]$. That aggregate includes tasks
permitting readback and is not a capability certificate. Masking, injection
and multi-seed checks test dependence on typed-neighbour evidence,
supporting structural evidence use on the tested families rather than
unrestricted materials reasoning (Appendix~\ref{app:limitations}).\\[3pt]
\noindent\ding{182}~We cast reasoning versus recitation as a controlled
representation test. An eight-view benchmark over seven task families varies
structural evidence through matched interventions, anti-shortcut checks and
a paired certification rule, holding question, gold answer and scaffold
fixed (Section~\ref{sec:method}). What makes those interventions possible is GraphSpace itself: it names each
structural relation separately, so a single orbit link or typed-neighbour
shell can be deleted or supplied while every other byte of the description
stays fixed, which is what makes the masked and injected arms comparable.\\
\noindent\ding{183}~We audit what produces the measured gain. Splitting the
headline by whether both views carry the required inputs separates a matched
representation effect from families whose plain rendering cannot answer at all. A solvability check on the two inputs gates every view comparison on that
distinction before the numbers are read, so a gap that only records one view's inability
to answer never enters a representation claim
(Section~\ref{sec:experiments}, Appendix~\ref{app:view-parity}).\\
\noindent\ding{184}~We turn the framework on its own benchmark. A rule that ignores the sibling link solved four hardened families, so we rebuilt
the sibling construction until eleven measured shortcut rules sit at chance.
On the rebuilt families the frozen model cites the pointer without using it,
while matched supervision makes it load bearing: following the link and then
reading the list it names is learnable, not absent
(Section~\ref{sec:mechanism}, Appendices~\ref{app:pointer-forced}
and~\ref{app:pointer-forced-sft}).

\section{Method}
\label{sec:method}

\begin{figure}[!t]
\centering
\includegraphics[width=\textwidth]{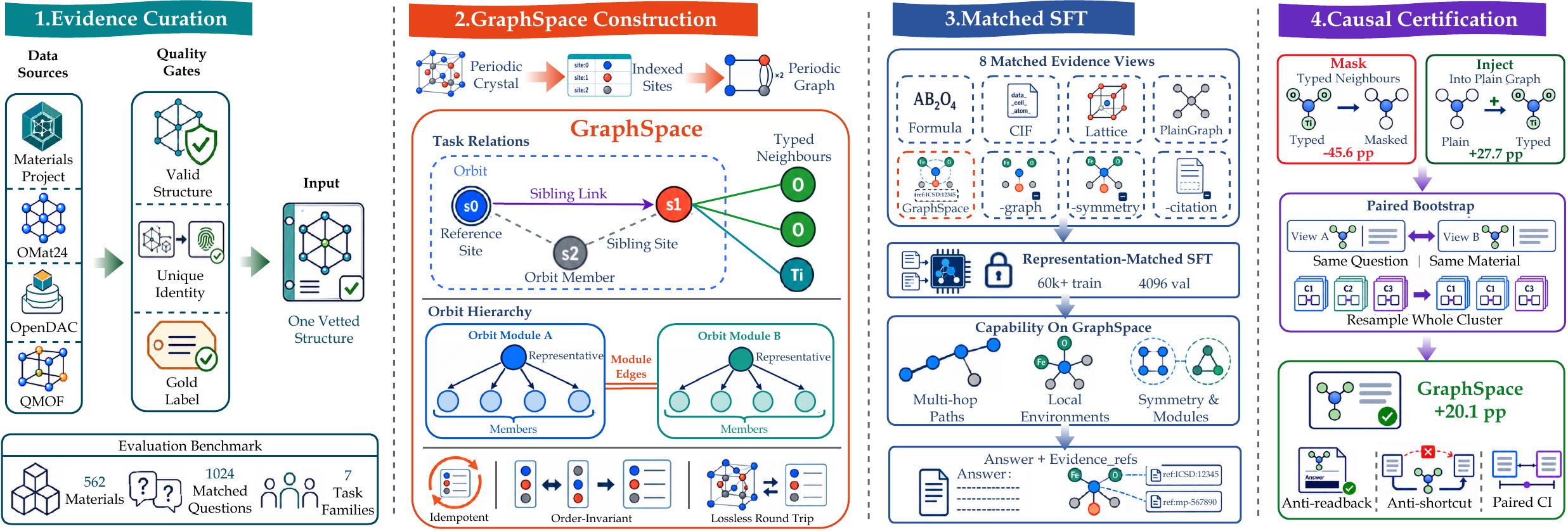}
\caption{\textbf{The CARAT framework.} Evidence curation, GraphSpace construction, matched SFT, and causal certification connect controlled inputs to scoped capability claims.}
\label{fig:carat-framework}
\end{figure}

Figure~\ref{fig:carat-framework} organizes CARAT into four stages.
\textbf{Evidence Curation} admits structures with valid geometry, unique
material identities and available gold labels. \textbf{GraphSpace
Construction} turns each vetted structure into explicit site, neighbour and
orbit relations. \textbf{Matched SFT} renders eight evidence views and
trains them under a shared schedule. \textbf{Causal Certification} tests the
answers through evidence interventions and material-paired statistics.
Across all four the material, question and gold answer stay fixed; the controlled variable is the structural evidence the model sees, which is
what makes a view comparison a statement about representation.

\subsection{Evidence Curation}
\label{sec:data}

One vetted material pool feeds every downstream representation. Materials
come from four public sources \citep{jain2013materialsproject}, and each
takes an identity from its atomic composition, space group and Wyckoff-site
pattern. That identity fixes its train, validation or test partition across
all views, so no other rendering of a training material can reach the test
set, and it is also the deduplication key at admission. Structures and
symmetry labels are computed deterministically with pymatgen
\citep{ong2013pymatgen} and spglib \citep{togo2024spglib}
(Appendix~\ref{app:uuid-ledger}). The result is $1{,}024$ matched questions
over $562$ materials in seven task families: CIF-plan reconstruction,
multi-hop EFS path reasoning, evidence sufficiency, graph construction,
local-environment reasoning, MOF module-property reasoning and symmetry
reasoning. Every admitted question is rendered under every view below, so
view comparisons share one evaluation target rather than drawing from
different question pools, and no view can quietly drop a question it cannot
carry.

\subsection{GraphSpace Construction}
\label{sec:gs-construction}

GraphSpace expands the vetted structure into a periodic graph and indexed
sites, and encodes the relations CARAT intervenes on as canonical text
built from four units: an \emph{orbit} identifies a Wyckoff position, a
\emph{site reference} identifies an atomic position within an orbit, a \emph{sibling} link (\texttt{orbit\_link}) connects same-orbit neighbours,
and a \emph{typed-neighbours} list records the element-tagged near-neighbour
shell of a site. The sibling link is the \emph{pointer} our later tests
delete: it says which neighbour list to read. Those four units are what make the interventions possible:
a relation, or the reference naming it, can be removed without replacing
the rest of the structural description.

Canonicalisation orders fractional coordinates, compresses parallel edges
and clamps the unit-cell boundary, and a hierarchical extension represents
module-level structure in large systems such as MOFs. The representation is idempotent, order-invariant, and faithful to the
audited fields at their written precision. \S\ref{sec:formalization} gives
the audits behind that claim, and Appendix~\ref{app:formal-details} marks
the boundary at which each of those audits stops.

The composer exposes that construction to the model as four external
references, one full GraphSpace view and three targeted ablations.
(1)~\textbf{Formula} contains only the reduced chemical formula;
(2)~\textbf{CIF} uses the raw crystallographic information file;
(3)~\textbf{lattice+sites} contains lattice vectors and fractional coordinates;
and (4)~\textbf{plain graph} contains periodic nodes and edges without
annotations.
(5)~\textbf{GraphSpace (full)} exposes the periodic graph, symmetry,
coordination environments, dimensionality, and evidence citations.
Its ablations remove (6)~the \textbf{graph} block, (7)~the \textbf{symmetry}
and local-motif blocks, or (8)~the \textbf{citation} markers.
The external references compare alternative ways of presenting the same
structure, and the three ablations isolate what each individual GraphSpace
evidence block contributes to the answer.

Let $q$ denote a question about a material, $y^\star(q)$ its gold answer, and
$v\in\mathcal{V}$ one of the eight views. The input is
\begin{equation}
x_{q,v} =
\bigl(\text{scaffold}(q),\,r_v(\text{struct}(q))\bigr),
\label{eq:carat-view-input}
\end{equation}
where $r_v$ renders the material structure into the selected evidence view.
The composer byte-copies the system prompt and task scaffold, preserves the
gold answer in the evaluation record and replaces only the structural
evidence block. Every view is checksummed, and admission requires identical
scaffolds and gold answers across all eight renderings, so the task and its difficulty stay fixed while only the available evidence
changes, and no incidental difference in prompt wording or in the answer key can enter or confound the comparison.

\subsection{Matched SFT}
\label{sec:training}

Every view receives the same base model, task coverage and optimization
schedule under Eq.~\ref{eq:carat-view-input}: full-parameter SFT of the 27B
model for $3$ epochs on $65{,}536$ training rows per view across all seven
task families, with a $4{,}096$-row matched validation set. Holding the schedule fixed makes the representation the controlled variable,
and the single material-identity hash means each of the $1{,}024$ test
questions appears once per view. Repeating all eight runs with seeds $42$,
$43$ and $44$ gives the $24$-arm matrix behind Table~\ref{tab:overall};
hyperparameters and archived training YAMLs are in
Appendix~\ref{app:training-hparams}.

\subsection{Causal Certification}
\label{sec:eval-protocol}

Certification turns the model outputs into tests of evidence dependence.
Masking removes typed-neighbour evidence from GraphSpace and injection adds
the corresponding evidence to a plain-graph input, with question and gold
answer unchanged, so the two interventions measure the loss of evidence and
the benefit of supplying it. Paired bootstrap then compares the same questions and materials, resampling
whole material clusters for the certification tests, and the verdict is
conditional on anti-readback, anti-shortcut and paired-confidence checks
rather than on an accuracy threshold alone. A mathematical property
of GraphSpace does not establish that a model uses it, just as a high score
does not establish the task is free of readback. Appendix
Table~\ref{tab:carat-gates} organizes the full battery by audit purpose and
by the boundary each pass licenses.

\begin{table}[!t]
\centering
\small
\setlength{\tabcolsep}{3.4pt}
\renewcommand{\arraystretch}{1.10}
\setlength{\abovecaptionskip}{3pt}
\setlength{\belowcaptionskip}{3pt}
\caption{\textbf{Per-view accuracy by task family.} 27B model, 1,024 matched
questions, seed 42, strict answer-exact rate; best per column bold.
Table~\ref{tab:family} gives the macro-hard-slice accuracy.}
\label{tab:overall}
\begin{tabular}{lccccccccc}
\toprule
& \textbf{Overall} & \multicolumn{7}{c}{\textbf{Per-family answer-exact rate}} \\
\cmidrule(lr){2-2}\cmidrule(lr){3-9}
\textbf{View} & \textbf{Micro} & \textbf{CIF plan} & \textbf{EFS path} &
\textbf{Suff.} & \textbf{Graph} & \textbf{Local} & \textbf{MOF} & \textbf{Symm.} \\
\midrule
\rowcolor{HeaderColor}
\multicolumn{9}{@{}l}{\textbf{Answer printed} (the answer field is verbatim in the evidence)} \\
GraphSpace full & \textbf{0.956} & \textbf{1.000} & 0.531 & \textbf{1.000} & \textbf{1.000} & \textbf{1.000} & \textbf{1.000} & \textbf{1.000} \\
\quad $-$~citations & 0.876 & \textbf{1.000} & 0.573 & \textbf{1.000} & 0.578 & \textbf{1.000} & \textbf{1.000} & \textbf{1.000} \\
\quad $-$~graph & 0.805 & \textbf{1.000} & 0.479 & \textbf{1.000} & 0.265 & \textbf{1.000} & \textbf{1.000} & \textbf{1.000} \\
\rowcolor{HeaderColor}
\multicolumn{9}{@{}l}{\textbf{Answer masked} (symmetry-local block removed)} \\
\quad $-$~symmetry & 0.922 & \textbf{1.000} & 0.562 & \textbf{1.000} & \textbf{1.000} & \textbf{1.000} & \textbf{1.000} & 0.712 \\
\rowcolor{HeaderColor}
\multicolumn{9}{@{}l}{\textbf{Alternative representations}} \\
Raw CIF & 0.761 & \textbf{1.000} & \textbf{0.708} & \textbf{1.000} & 0.289 & \textbf{1.000} & 0.618 & 0.848 \\
Plain graph & 0.821 & \textbf{1.000} & 0.667 & \textbf{1.000} & \textbf{1.000} & \textbf{1.000} & 0.331 & 0.545 \\
Lattice + sites & 0.682 & \textbf{1.000} & 0.615 & \textbf{1.000} & 0.255 & \textbf{1.000} & 0.316 & 0.667 \\
Formula & 0.542 & 0.649 & 0.604 & \textbf{1.000} & 0.162 & 0.739 & 0.316 & 0.394 \\
\bottomrule
\vspace{-17pt}
\end{tabular}
\end{table}

\textbf{Matched comparisons.}
The evaluation unit is a question about one material, paired across views by
\texttt{eval\_id}: for inputs of the form in
Eq.~\ref{eq:carat-view-input} the gold answer $y^\star(q)$ and scaffold are
shared and only $r_v$ changes, so accuracy differences compare responses to
different evidence for the same task rather than performance on different
questions. With the matched training schedule, that is what controls the
representation contrast. Whether the contrast reflects structural evidence
use is a separate question, and it is the one the shortcut and intervention
diagnostics test.

\textbf{Anti-shortcut controls.}
CARAT tests three ways in which a correct answer can fail to demonstrate the
intended capability: \textbf{recitation}, where a view prints the gold
answer and the response may be copied rather than derived;
\textbf{tolerance chasing}, where a symmetry label tracks the
\texttt{symprec} used to generate it; and \textbf{pattern matching}, where
responses follow surface tokens rather than the structural change a
counterfactual edit makes. Capability is credited only on answer-absent
views. Appendix~\ref{app:diagnostics} gives each diagnostic and the
memorization controls, none of which shows a measurable benefit from the
tested cues (Appendix Table~\ref{tab:memorization}).

\setlength{\intextsep}{4pt}

\textbf{Paired inference and certification.}
\label{sec:stats}
Views are compared question by question rather than as independent groups,
and each analysis reports its registered resampling unit and estimator:
material-cluster analyses resample whole materials to keep their questions
together, and the input-intervention comparisons use paired case bootstrap,
which is not interchangeable with the material-level estimates
(Appendix~\ref{app:metric-defs}). Capability comparisons account for chance
baselines computed from the marginal answer distribution, and the Holm
correction \citep{holm1979simple} holds the family-level error rate when
several comparisons are read together.

The broader benchmark's macro-hard-slice accuracy measures aggregate
representation performance and retains readback-eligible tasks. Capability claims instead use the readback-hardened families, or
\emph{healthy pool}, admitted by two gates: anti-readback excludes families
whose gold answer is printed verbatim in the input, and the null baseline
compares family scores with chance computed from the marginal answer
distribution. A scoped claim then requires a paired gain on that admitted
pool with a $95\%$ confidence interval strictly above zero, and the verdict
it earns applies to the tested representation contrast and admitted task
pool. Every reported gain is therefore tied to a matched comparison, an eligible
task pool and a recomputed uncertainty estimate
(Appendix~\ref{app:uuid-ledger}).

\section{Experiments}
\label{sec:experiments}

\textbf{Setup.} Two evaluation sets answer different questions. The matched
benchmark is 1,024 questions over 562 materials in the eight views of
\S\ref{sec:training}. The readback-hardened pool is larger and stricter:
77,514 questions over seven families from held-out materials, the answer
token withheld from both views by construction and no question shared with
the benchmark. Representation-utility claims use the first set, capability
claims the second (\S\ref{sec:eval-protocol}).

Table~\ref{tab:overall} gives the strict answer-exact rate for all eight
views: grounded GraphSpace leads at $0.956$ against Formula $0.542$. Under the macro-hard-slice metric the gate uses, the paired GraphSpace
versus formula difference on seed~42 is $+0.173$, 95\% CI $[+0.158,
+0.187]$. The two figures differ because the macro metric gives partial
credit within an item and weights the six hard slices equally, which lifts the formula view from $0.542$ to $0.783$ while leaving
GraphSpace at $0.956$. The slices regroup the seven families and each
carries the same weight whatever its size, so a $72$-item slice counts as
much as a $266$-item one (Appendix~\ref{app:metric-defs}).

\begin{table}[!tb]
\centering
\small
\setlength{\tabcolsep}{4pt}
\renewcommand{\arraystretch}{1.14}
\setlength{\abovecaptionskip}{3pt}
\setlength{\belowcaptionskip}{3pt}
\caption{\textbf{Multi-seed certification matrix.} Seeds 42, 43 and 44 on
the same paired set of 1,024 questions. The EFS-path family breaches the
two-point tolerance in every seed.}
\label{tab:family}
{\setlength{\aboverulesep}{0pt}\setlength{\belowrulesep}{0pt}%
\begin{tabular}{crrrlrl}
\toprule
\rowcolor{HeaderColor}
\textbf{Seed} & \textbf{GS full} & \textbf{Best-comp.} & \textbf{Formula} &
\textbf{$\Delta$ vs formula (95\% CI, pp)} & \textbf{$\Delta$ worst (pp)} &
\textbf{Worst family} \\
\midrule
42 & \textbf{0.956} & 0.944 & 0.783 & $+17.3\;[+15.8,+18.7]$ & $-8.9$ & EFS path \\
43 & \textbf{0.958} & 0.943 & 0.789 & $+16.9\;[+15.5,+18.3]$ & $-5.7$ & EFS path \\
44 & \textbf{0.960} & 0.943 & 0.780 & $+18.0\;[+16.5,+19.5]$ & $-8.9$ & EFS path \\
\bottomrule
\end{tabular}}
\end{table}

\begin{figure}[!t]
\centering
\includegraphics[width=0.88\linewidth]{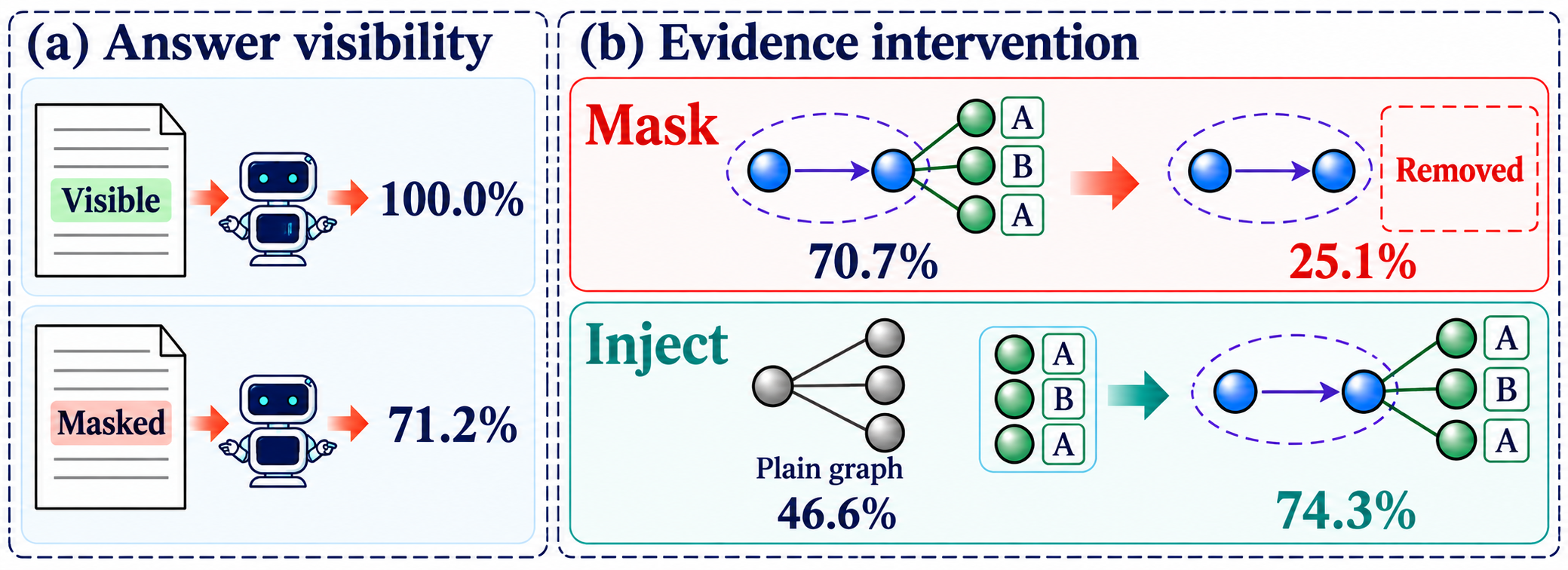}
\caption{\textbf{Answer visibility and evidence intervention.} Each intervention uses its own paired cohort; full uncertainty estimates appear in Appendix Table~\ref{tab:A1}.}
\vspace{-5pt}
\label{fig:overview}
\end{figure}

A second, larger measurement holds the checkpoint fixed and varies only the
input view. On the hardened pool, rendering the same questions as
GraphSpace rather than as a plain graph is worth $+0.193$, 95\% CI
$[+0.191,+0.195]$. Retraining on each view separates the representation
effect from the training format: GraphSpace wins on \textbf{all eight
training views}, from $+0.136$ trained on raw CIF to $+0.211$ trained on
formula text, every bootstrap interval excluding zero. The gain therefore tracks the evidence the input carries at inference, not
the training format (Appendix~\ref{app:healthy-pool}); \S\ref{sec:mechanism}
takes up what the size of that gap licenses.

Figure~\ref{fig:overview} separates answer visibility from causal dependence
on typed evidence: panel (a) contrasts full and symmetry/local-masked
GraphSpace views, panel (b) removes and supplies typed evidence at a fixed
checkpoint (\hyperref[sec:A1]{the mask-and-inject analysis}), and a length
control pads the plain-graph input without closing the gap
(Appendices~\ref{app:readback} and~\ref{app:scope-limits}). The three
GraphSpace ablations then localize each block: removing the periodic graph
is the largest single ablation on the benchmark aggregate ($0.956 \to
0.899$) and is concentrated in graph-sensitive families, so the graph block
carries information the model uses, while the symmetry/local and citation
blocks cost less. Ablations show \emph{that} a block matters, not \emph{how} it is used, which
\S\ref{sec:mechanism} tests directly.

\section{Analysis}
\label{sec:analysis}

Section~\ref{sec:experiments} establishes \emph{that} the grounded view
helps and by how much. The three subsections below ask why. Each puts one part of the claim under the same tests: the model in
\S\ref{sec:mechanism}, the representation in \S\ref{sec:formalization}, the
evaluation in \S\ref{sec:certification-analysis}.

\begin{table}[!b]
\centering
\caption{\textbf{Citing a relation versus using it.} Held-out
pointer-forced families, GraphSpace view. Shortcut maximum $27.0\%$;
Appendix~\ref{app:pointer-forced-sft} gives the denominators and reads the
twin-agreement row, over pairs that differ only in the pointer.}
\label{tab:pointer-main}
\begin{tabular}{llrrr}
\toprule
\rowcolor{HeaderColor}
\textbf{Condition} & \textbf{What it tests} & \textbf{$n$} & \textbf{Frozen} & \textbf{Pointer-SFT} \\
\midrule
$k{=}4$, pointer shown   & selection $+$ computation & 882   & 32.3\% & \textbf{99.8\%} \\
$k{=}4$, pointer deleted & residual shortcut         & 882   & 26.8\% & 23.4\% \\
Pointer moved (twins)    & selection alone           & 1{,}764 & 25.9\% & \textbf{99.9\%} \\
Single list (control)    & computation alone         & 882   & 60.2\% & 99.8\% \\
\midrule
\multicolumn{2}{l}{Twin agreement when only the pointer moves} & 882 & 95.6\% & \textbf{0.11\%} \\
\multicolumn{2}{l}{Paired effect of showing the pointer (pp)}  & 882 & $+5.56$ & $\mathbf{+76.42}$ \\
\bottomrule
\end{tabular}
\end{table}

\begin{figure}[!t]
\centering
\includegraphics[width=0.78\linewidth,trim=0 20 0 12,clip]{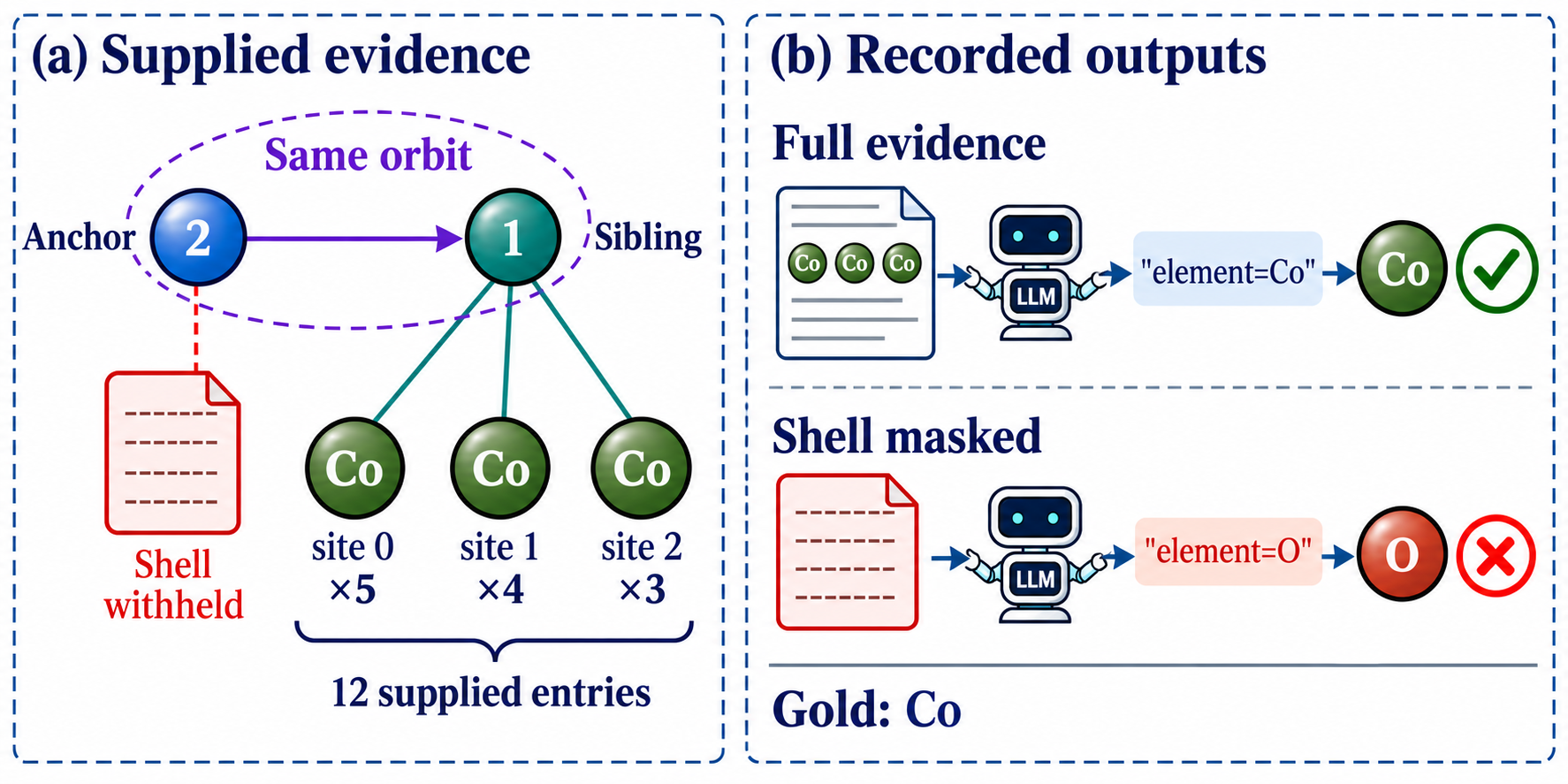}
\caption{\textbf{Grounded citation and unsupported completion in a held-out case.} The schematic contrasts supplied Co neighbours with an unsupported O completion after masking.}
\vspace{-4pt}
\label{fig:thinking-evidence}
\end{figure}

\begin{figure}[!b]
\centering
\includegraphics[width=0.85\linewidth]{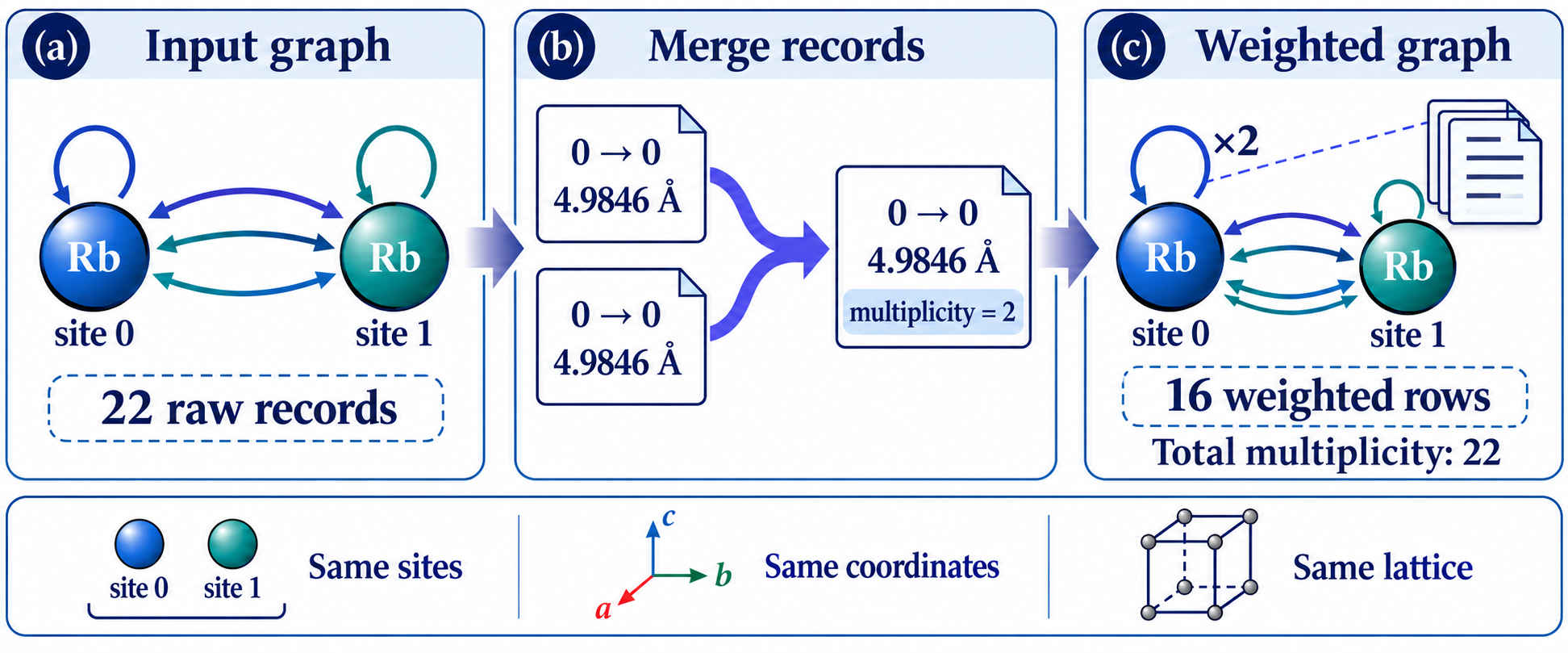}
\vspace{-2pt}
\caption{\textbf{Canonicalization of an archived Rb record.} Schematic graphs illustrate duplicate-edge merging while preserving multiplicity, sites, coordinates, and lattice.}
\label{fig:canonical-example}
\end{figure}

\subsection{Does the model reason?}
\label{sec:mechanism}

CARAT's view comparisons localize \emph{that} the grounded view helps, not
how. The largest gap is the wrong place to look for it. That gap falls on the
\texttt{S4c}-type ``anchor's dominant neighbour element via sibling'' task,
whose plain view carries neither the pointer nor the element labels the
question needs. Where both views carry the same inputs the gap is
$+1.96$~pp, an $n$-weighted mean over the four matched families
(\S\ref{sec:experiments}, Appendix~\ref{app:view-parity}). Two
input-side tests then converge on the same field. Correlationally,
grounded cases that cite both an \texttt{orbit\_link} and a \texttt{sibling}
literal answer S4c correctly $72.1\%$ of the time, against $20.0\%$ for
cases citing neither,
while in plain graph the citation rate is zero because the field does not
exist there. Causally, deleting the typed-neighbour shell on top of the pointer fields
costs $-45.6$~pp against the unmasked baseline, and injecting the same typed
edges into plain graph adds $+27.7$~pp, both intervals excluding zero
(Figure~\ref{fig:thinking-evidence}). The two legs move in opposite directions on the same field, and the additive
leg attributes the gap to what the representation contains. Training the model to emit an
explicit thinking trace sharpens the same signature, worth a paired IO gain
of $+3.16$~pp on the grounded view with three preregistered alignment gates
passing at the strong tier. Every behavioural layer agrees in sign, and what
they all track is the element-tagged typed shell, not the pointer fields naming it (\S\ref{app:A0}--\S\ref{app:A3}, with the
full protocol in Appendix~\ref{app:mechanism-details}).

\begin{figure}[!tb]
\centering
\includegraphics[width=\linewidth]{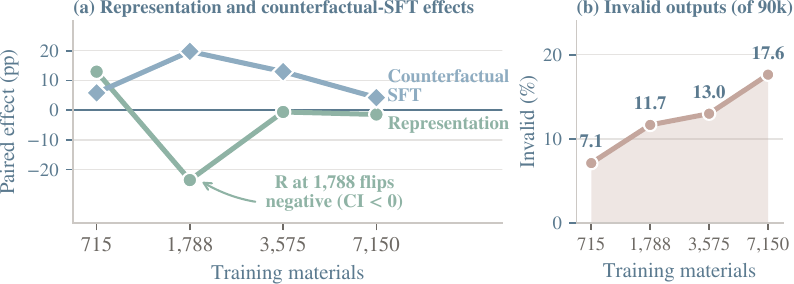}
\caption{\textbf{More training data does not stabilize the measured effects.} Panel (a) shows material-cluster 95\% intervals; panel (b) pools 90,000 predictions per training size.}
\label{fig:scaling-story}
\end{figure}

\textbf{Forcing the composition.} None of that separates reading a relation
from reading a list, because in the released families one typed neighbour
list is the only candidate, so following the pointer is never needed to find
it (Appendix~\ref{app:pointer-blind}). Separating the two needs several candidate lists carrying different answers,
so that the pointer is the only way to choose. We rebuilt the sibling
construction that way: each case offers $k{=}4$ candidate neighbour shells
whose first-shell answers are pairwise distinct, so \texttt{orbit\_link}
alone identifies the right one. The other shortcut-prone families are not
separately rebuilt here. Eleven pointer-blind rules, written against our own construction, top out at
$27.0\%$ on the rebuilt set, and Table~\ref{tab:pointer-main} reads the four
conditions the construction affords. Frozen, the checkpoint answers
the unambiguous control well enough to show it can do the computation and
transcribes the pointer target wherever it reports one, yet deleting that pointer barely moves it: it returns the same answer for
$95.6\%$ of the pairs in which only the pointer moved, citing the relation
without using it to select the evidence. After pointer-forced supervision every condition separates: with the pointer
shown the model reaches $99.8\%$, over three times the best of the eleven rules, and deleting the pointer
collapses it to $23.4\%$, back inside the band those rules occupy. Deletion
leaves a pointer-blind rule untouched, so none can account for the $99.8\%$. The composition is learnable here, and the frozen model's own use of the
pointer is weak rather than impossible: worth $+5.56$~pp on the same
evidence
(Appendices~\ref{app:pointer-forced} and~\ref{app:pointer-forced-sft}).

\begin{figure}[!tb]
\centering
\includegraphics[width=\linewidth]{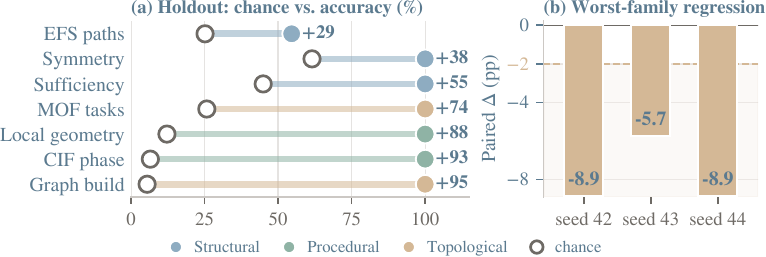}
\caption{\textbf{Generalization and the limits of certification.} (a) Fluorine-holdout accuracy, material-cluster 95\% intervals. (b) Worst-family regression per seed; all three breach the $-2$~pp tolerance.}
\vspace{-4pt}
\label{fig:scope-limits}
\end{figure}

\subsection{Formal properties of GraphSpace}
\label{sec:formalization}

If the advantage lives in the representation, the representation itself
must be well specified (Figure~\ref{fig:canonical-example}): canonical,
information-conserving, forward-extensible and consistent across materials.
Table~\ref{tab:graphspace-properties} summarizes the five verdicts, all
read-only on frozen release bytes and altering no main-experiment number.
Each row names the check that was run, not a general property:
\emph{Scalability} is a pass rate on $170{,}290$ hierarchical rows, not a time or memory measurement; \emph{Cross-material} is $100\%$
on symmetry variants and synthetic supercells, with elemental substitution
at $82.5\%$ (Appendix~\ref{app:B8}), where seven exceptions track
species-dependent bond lengths.

\Needspace*{10\baselineskip}
\begin{wraptable}[9]{r}{0.48\textwidth}
\centering
\small
\setlength{\tabcolsep}{4pt}
\renewcommand{\arraystretch}{1.15}
\setlength{\abovecaptionskip}{2pt}
\setlength{\belowcaptionskip}{2pt}
\caption{\textbf{GraphSpace audits.}}
\label{tab:graphspace-properties}
{\setlength{\aboverulesep}{0pt}\setlength{\belowrulesep}{0pt}%
\begin{tabular}{lrl}
\toprule
\rowcolor{HeaderColor}
\textbf{Property}    & \textbf{Result}      & \textbf{Verdict} \\
\midrule
Canonical order      & $14{,}262$ / $14{,}262$ & Pass  \\
Round-trip decode    & $14{,}262$ / $14{,}262$ & Pass  \\
Scalability          & $170{,}290$ / $170{,}290$ & Pass \\
Cross-material       & $100\%$              & Pass     \\
No redundancy        & $+11.8$ pp           & Strong   \\
\bottomrule
\end{tabular}}
\end{wraptable}

\textbf{What the audits establish.} The canonical form is a deterministic
serialisation: a three-key site order, edge grouping by $(\text{src},
\text{dst}, \text{round}_4(d), \{e_i, e_j\})$ with an explicit
\texttt{multiplicity} field, and a periodic-boundary clamp. Idempotency,
information conservation and sort-key unambiguity hold on every audited row
of both the flat form and the hierarchical refinement, with no collision in $184{,}552$ rows. GraphSpace is therefore lossless
for the fields it stores, up to the $10^{-4}$~\AA{} quantisation
$\mathrm{round}_4$ discards, on every release we audited, and training on
the canonical form rather than the expanded one is worth $+11.77$~pp on the frozen hardened pool.

\begin{figure}[!tb]
\centering
\includegraphics[width=\linewidth]{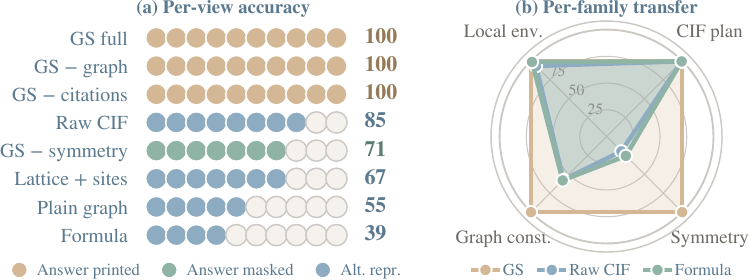}
\caption{\textbf{Answer visibility across views and external transfer.} (a) Per-view accuracy by answer-visibility cohort. (b) External-benchmark macro accuracy for four task families under three views.}
\label{fig:readback-controls}
\end{figure}

The formal and behavioural lines then meet at counterexample CE-2
(Appendix~\ref{app:B7}), which names exactly the fields whose removal and reinstatement produce the mask
and injection effects of \S\ref{sec:mechanism}: the property that makes the
representation well specified is the property the model's accuracy depends
on. Appendix~\ref{app:B7} pairs each lemma with its bounding counterexample;
\S\ref{app:B1B2}--\S\ref{app:B9} carry the audit tables, the \texttt{spglib}
standardisation the serialisation is relative to, and the
elemental-substitution and edge-case results for every audited release.

\begin{figure}[!tb]
\centering
\includegraphics[width=\linewidth]{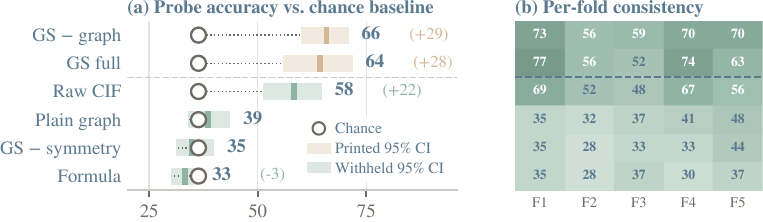}%
\vspace{-4pt}
\caption{\textbf{Layer-48 space-group probe accuracy by answer visibility.} (a) probe mean and 95\% CI vs.\ the majority baseline; (b) per-fold accuracy. Rows above the dashed rule print the answer.}
\vspace{-4pt}
\label{fig:probe-visibility}
\end{figure}

\phantomsection\label{sec:formalization-unified}
\subsection{What does the protocol certify?}
\label{sec:certification-analysis}

We now ask what the evaluation licenses, running CARAT's decision rule
against training scale, a structural holdout, three seeds and the
answer-visibility cohorts (Appendix~\ref{app:limitations}).

\textbf{The rule admits a scoped gain and refuses a global one.}
Figure~\ref{fig:scaling-story} compares four training sizes on one fixed
protocol: the representation effect changes sign and the supervision effect
peaks at an intermediate size, so scale is not a monotonic remedy.
Figure~\ref{fig:scope-limits}(a) adds a fluorine holdout on which EFS path
clears its majority baseline. Across three seeds
(Figure~\ref{fig:scope-limits}(b)) GraphSpace leads on the aggregate point
estimate and on the paired difference from formula inputs in every seed,
while EFS-path regression breaches the $-2$~pp tolerance in all three. The rule therefore admits the healthy-pool gain as a scoped claim and
withholds the global one, exactly as designed.

\textbf{Answer visibility drives measured accuracy.} In Figure~\ref{fig:readback-controls} the three answer-printing views sit
above every alternative representation, and removing the symmetry/local
block costs $28.8$ points. The same ordering appears on the external benchmark and in the layer-48
probe (Figure~\ref{fig:probe-visibility}). These scores track label
visibility, so CARAT reports those families as diagnostics, not capability.

\section{Related work}
\label{sec:related}

\textbf{Representations and prediction for crystalline materials.}
Crystal-graph networks learn property predictors from periodic structures
\citep{xie2018cgcnn,chen2019megnet,choudhary2021alignn}, MACE learns
interatomic potentials \citep{batatia2022mace}, and text serves property
prediction, crystal tokenization, predictive chemistry and invariant string
encodings
\citep{rubungo2025llmprop,gruver2024crystalllm,jablonka2024gpt,xiao2023slices},
with LLM4Mat-Bench comparing composition, CIF and text
\citep{rubungo2025llm4mat}. None of those objectives identifies which
structural information supports an answer. Rel-LLM \citep{wu2025relllm} raises the same concern for relational
databases, where flat serialisation hides the key structure. CARAT fixes the question and gold answer, intervenes on evidence
fields and excludes readback, turning a stated readiness concern \citep{miret2025llms} into a test: does
the answer survive deleting what the input prints? 

\textbf{Shortcut learning and reasoning faithfulness.}
Predictive success can depend on cues other than the intended mechanism
\citep{geirhos2020shortcut}, meaning-preserving prompt changes move measured performance
\citep{sclar2024prompt,mizrahi2024multiprompt}, and a coherent chain of
thought need not describe the computation behind an answer
\citep{turpin2023cot,lanham2023measuring,dziri2023faith}. CARAT targets that case directly: wording stays fixed while interventions
vary the evidence.

\textbf{Interventions and reliable evaluation.}
CheckList organizes behavioural evaluation around capabilities and test
types \citep{ribeiro2020checklist}, while causal tracing, model editing and
causal abstraction relate internal computation to behaviour
\citep{meng2022rome,geiger2021causalabstraction}. CARAT works at the representation level through matched mask-and-inject
tests, separating whether an answer depends on a field from where it is
computed. Selective prediction and calibration inform the refusal
diagnostics \citep{geifman2017selective,guo2017calibration}, and bootstrap
inference with Holm correction \citep{efron1993bootstrap,holm1979simple}
supplies the uncertainty and the multiple-comparison control that every gain
reported here carries.

\section{Conclusion}

CARAT decides whether a materials LLM reasons over crystal structure or
recites an answer printed in its input. On a 27B model and a
$1{,}024$-question benchmark, representation ablation and an anti-shortcut
protocol certify a scoped advantage for typed, element-labelled evidence on
readback-hardened families. We then audited it twice. Split by whether both views can answer, the
headline $\Delta = +0.193$ is $+1.96$~pp where they can and $+46.7$~pp where
they cannot, so most of it records a missing input. And a rule that ignored
the sibling link answered four of seven hardened families, so we rebuilt
that construction until the link mattered: matched supervision reaches
$99.8\%$, and deleting the link drops it below the best shortcut. Following
the link and reading the list it names is learnable here. A score is not a capability until someone has tried
a shorter route, and a view comparison is not a representation result until
both views answer (Appendix~\ref{app:consolidated}).

\subsubsection*{Ethics statement}
This work involves no human subjects, no personally identifying data,
and no user study. All structures come from public materials databases
(Materials Project, QMOF, CoRE MOF, OpenDAC) and are used under their
published terms; no proprietary or restricted structure data enters any
release. The evaluation releases are diagnostic: each manifest sets
\texttt{training\_allowed} to false, so a release used for evaluation
cannot be folded back into training and inflate a later result.

The main foreseeable harm from this line of work is scientific rather
than social: a benchmark that rewards answer recitation would license
overconfident claims about model capability in a domain where
downstream decisions are experimental and costly. The paper is built
against that failure. We report the families where certification is not
granted, the seeds where the tolerance is breached, and the mechanistic
contrast whose interval includes zero, and we state in
Section~\ref{sec:mechanism} which alternative explanation our
experiments cannot exclude. We claim no chemical validity for any
generated candidate and no capability beyond the tested families. We
adhere to the ICLR Code of Ethics.

\subsubsection*{Reproducibility statement}
The benchmark is built by an additive composer that byte-copies the shared
scaffold and gold answer across all eight views and swaps only the structural
evidence block; every view is checksummed and a question is admitted only if its
eight views share an identical scaffold and gold answer
(Section~\ref{sec:data}). All splits are cut by material identity with zero
overlap. Analyses resample their stated unit, which is material clusters, hard-slice
strata, cases or folds, with Holm correction where specified.
The decision procedure recomputes and digit-checks every reported confidence
interval and includes an adversarial tamper test (Section~\ref{sec:stats});
that check runs against the result files, and keeping the manuscript in step
with them is a separate, manual step. Where we write \emph{preregistered},
the evidence is a rule written into the verdict log before its results,
which fixes the order of writing and not an external registration. Per-view and per-family scores, checksums, and the frozen query set
accompany the release. The supplementary archive is self-contained and needs
no network access and no model inference: it ships the matched
$1{,}024$-question benchmark in all eight views, a case-matched sample of
the readback-hardened pool with the full manifests, the rebuilt
pointer-forced pack, the canonical-form and scoring code, the verdict logs,
and three standard-library-only scripts that recompute the gold parity
across views, the $+0.193$ headline, and its $+1.96$ / $+46.7$~pp split.
Appendix~\ref{app:accounting} states the scoring formulas and their
denominators, the composition of both evaluation sets, and every rule
that removes an item from a reported denominator;
Appendix~\ref{app:training-hparams} names the base checkpoint and the
training schedule; Appendix~\ref{app:uuid-ledger} lists the frozen
release paths, manifest checksums, bootstrap seeds, verdict logs, and
the scorer and statistics entry points needed to recompute each
headline number.

\subsubsection*{AI use statement}
Generative AI was used in two roles, both disclosed here. First, as a
writing aid: drafting and editing prose and \LaTeX{} for this manuscript.
Second, in benchmark construction, a language model paraphrased question
templates into natural-language variants; it changed only the wording of a
question, never the gold answer, the structural evidence, or the scorer.
Both roles are bounded by the pipeline itself. Gold answers and symmetry
labels are computed deterministically with pymatgen and spglib, every
evidence view is rendered by our own composer from those computed fields,
and the scorer is frozen. The paper also measures what the wording is worth:
crossing training and evaluation phrasing leaves off-diagonal accuracy at
$96.20$--$99.44\%$, and surface edits to order, whitespace and names move
accuracy by at most $0.62$~pp (Appendix~\ref{app:readback}), so no result
reported here rests on a model-authored phrasing. No model contributed
research ideas, experimental design, analysis, or any reported number: every
number was produced by the authors' evaluation pipeline and checked against
on-disk records. All AI-assisted text was reviewed and revised by the
authors, and the authors take full responsibility for the content, claims,
and artifacts of this paper.

\bibliography{iclr2027_conference}
\bibliographystyle{iclr2027_conference}

\clearpage
\appendix
\setlength{\intextsep}{4pt}

The appendix backs up the main text in nine blocks, in the order a reader
checking a claim would want them. Appendix~\ref{app:limitations} states the
scope of every main-text claim. Appendix~\ref{app:analysis} collects the
additional diagnostic figures, grouped by the claim they test, and
Appendix~\ref{app:detailed-tables} carries the numerical tables the main
text summarises. The next two give the full protocols behind
Section~\ref{sec:analysis}: Appendix~\ref{app:mechanism-details} the
per-layer detail for the mechanism analysis, including the ASCII layout of
the reader circuit, and Appendix~\ref{app:formal-details} the audit protocol
for the formal properties, including the boundary counterexample table.
Appendix~\ref{app:accounting} defines every metric and accounts for how each
reported number is computed, and Appendix~\ref{app:consolidated} states what
the three analysis lines jointly establish and what would extend each of
them. The last two are for rerunning the work:
Appendix~\ref{app:training-hparams} lists the training hyperparameters
shared by every SFT arm, and Appendix~\ref{app:uuid-ledger} documents the
reproducibility artefacts.

\section{Limitations}
\label{app:limitations}

The main text reports what the controlled interventions establish and
Section~\ref{sec:certification-analysis} shows where that support stops. This
appendix
states the boundary conditions behind those figures.

\subsection{Training scale and the structural holdout}

The representation effect in Figure~\ref{fig:scaling-story} averages
GraphSpace-minus-plain contrasts over clean and counterfactual-supervised
training. The supervision effect averages counterfactual-minus-clean
contrasts over both representations. Neither curve is therefore an
SFT-versus-base comparison. Across the four measured sizes invalid outputs rise from $7.1\%$ to
$17.6\%$. These runs therefore bound what scale explains: they identify no
scaling law, and they do not attribute the parsing failures to scale.

On the fluorine holdout of Figure~\ref{fig:scope-limits}(a), EFS path
accuracy is $54.6\%$ with a 95\% confidence interval of $[45.7,64.0]$,
against a majority baseline of $25.2\%$. Appendix~\ref{app:generalization}
gives the holdout's size, its composition and the rest of the
structural-holdout detail.

\subsection{The global decision rule}

The confidence-interval criterion displayed in
Figure~\ref{fig:scope-limits}(b) concerns the formula contrast rather than
every competing view. EFS path regression against the best competing view is
$-8.85$, $-5.73$ and $-8.85$~pp over the three seeds, breaching the $-2$~pp
tolerance each time.

\subsection{Probe accuracy by answer visibility}

Across the six views of Figure~\ref{fig:probe-visibility} the unweighted
probe mean is $41.2\%$ when the printed field is withheld and $65.1\%$ when
it is present. Those intervals resample fold accuracies rather than paired
material-level effects, so the difference is descriptive and cannot on its
own rule out recitation.

\subsection{The length control}
\label{app:scope-limits}

Padding the plain-graph input at the same checkpoint moves accuracy by
$+0.04$~pp, with a stratified 95\% confidence interval of $[-0.61,+0.67]$.
The residual GraphSpace contrast over the same cases is $+6.51$~pp, interval
$[+5.17,+7.85]$. Padding therefore does not explain the gap, although the
inputs are not exactly token-matched and the two view checkpoints are
separately trained.

\subsection{Mechanistic and reference boundaries}

Both SFT views improve deep-layer probe decodability, while the pooled
reader-versus-random ablation interval includes zero: a three-seed contrast
of $+1.82$~pp, interval $[-0.20,+3.98]$
(Appendix Fig.~\ref{fig:app-mechanisms}). The MACE-MP comparison (Appendix Fig.~\ref{fig:baselines}) is a
professional-tool reference rather than a paired representation test,
because its unique frame--task denominator differs from the repeated LLM
prompts. Four boundaries fix how far the rest carries. The eight training
views are eight renderings for one model family, so a view ordering is a
statement about that family. The pointer-forced result is one seed and one
construction, and it is a claim about the sibling composition rather than
about the other hardened families. Holding out materials tests transfer
across materials within a task family and not transfer across task families.
And the frozen union checkpoint and the pointer-forced arm have different
training histories, so the difference between them is not an attribution to
any single component of supervision. Broader tool comparisons and transfer across model scales remain future
work, and none of the claims above depends on either.

\clearpage
\section{Additional CARAT diagnostics}
\label{app:analysis}
The following panels group controls by the claim they test. Main-text
comparisons are not repeated; each group retains its statistical scope
and an interpretation of the result.

\label{app:readback}
\noindent\begin{minipage}{\linewidth}
\begin{figure}[H]
\centering
\includegraphics[width=\linewidth]{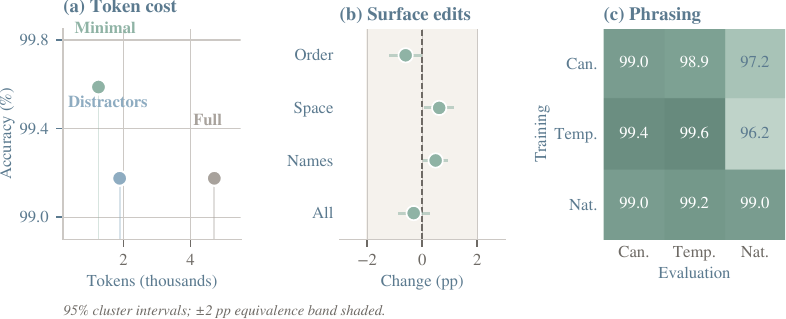}
\caption{\textbf{Token cost, surface edits, and query phrasing.} Style whiskers are material-cluster 95\% intervals. Shading marks the pre-specified two-point equivalence band.}
\label{fig:app-rendering}
\end{figure}
\smallskip
Figure~\ref{fig:app-rendering}(a) compares full, minimal, and
distractor-augmented schemas on 728 cases. Minimal facts reduce token
cost by 73.5\%, while the accuracy-difference interval includes zero.
Panel (b) changes order, whitespace, and names on 1,024 fixed questions;
point effects are at most 0.62~pp. Panel (c) crosses training and
evaluation phrasing: off-diagonal accuracy remains 96.20--99.44\%.
These controls support robustness to the tested rendering changes,
not arbitrary language variation.
\end{minipage}
\medskip

Panels~(a) and~(b) of this appendix's answer-visibility material are
displayed as Figures~\ref{fig:readback-controls}
and~\ref{fig:probe-visibility} in the main text. How to read them:
Figure~\ref{fig:readback-controls}(a) is a ten-percent pictogram
whose filled circles carry the cohort colour, and panel
(b) is a radar chart of seed-mean macro accuracy;
Figure~\ref{fig:probe-visibility}(a) marks the majority baseline with a
hollow marker, and panel (b) is a per-fold heatmap over those views. On the
external benchmark of Figure~\ref{fig:readback-controls}(b), GraphSpace
scores at ceiling where the answer is printed, while raw CIF and formula
inputs average in the 60--70\% range.

\noindent\begin{minipage}{\linewidth}
\begin{figure}[H]
\centering
\includegraphics[width=\linewidth]{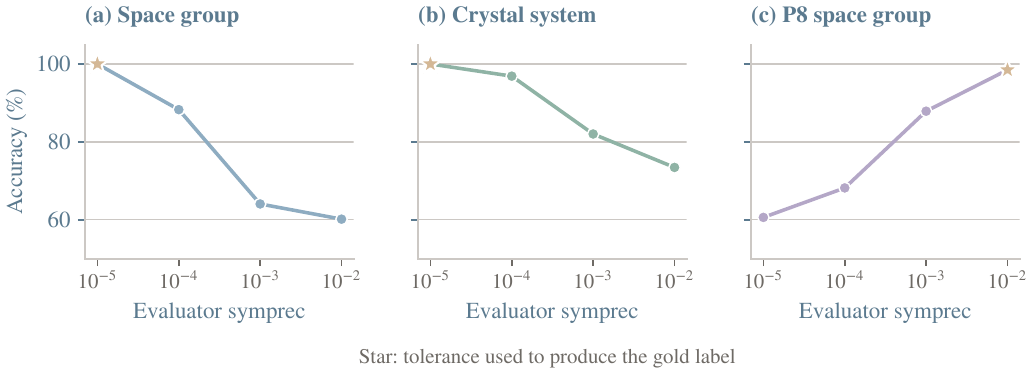}
\caption{\textbf{Evaluator symmetry-tolerance sensitivity.} Stars mark the gold-label tolerances.}
\label{fig:app-tolerance}
\end{figure}
\smallskip
Figure~\ref{fig:app-tolerance} finds that each accuracy curve peaks
at its gold-generation tolerance. This exposes coupling between the
generator and evaluator rather than independent physical reasoning;
the sweep is a methodological diagnostic excluded from capability scoring.
\end{minipage}
\medskip

\subsection{Generalization and mechanism}
\label{app:generalization}
\noindent\begin{minipage}{\linewidth}
\begin{figure}[H]
\centering
\includegraphics[width=\linewidth]{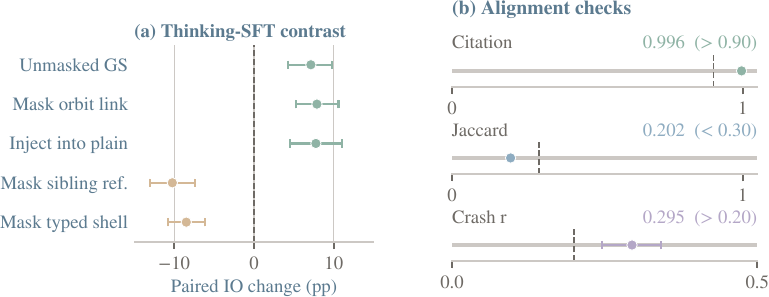}
\caption{\textbf{Thinking-SFT contrasts and alignment checks.} Whiskers show available 95\% intervals; dashed lines mark each check's own threshold.}
\label{fig:thinking-stats}
\label{fig:a2mask}
\label{fig:a26gates}
\end{figure}
\smallskip
The fluorine holdout behind Figure~\ref{fig:scope-limits}(a) is $5{,}404$
questions over $636$ materials under one training seed, and the
probe panels of Figure~\ref{fig:probe-visibility} use $132$ materials
per view with $500$ resamples of fold accuracies rather than paired
material-level effect intervals.
Figure~\ref{fig:thinking-stats}(a) compares each thinking-SFT view with its
own matched frozen baseline, not with the unmasked trained view. Its bars
are the predicted-only scoring, so they read $+7.13$, $-10.24$ and $-8.49$;
Table~\ref{tab:A2v6} and \S\ref{sec:mechanism} report the same contrasts
over the common denominator of $1{,}900$ with non-parsing counted wrong, at
$+3.16$, $-11.16$ and $-7.89$, and Appendix~\ref{app:paired-subsets} gives
both.
Panel (b) reports separate citation, cross-view overlap and citation--crash
checks, whose three scales are not a common reasoning score. The main-text
case illustrates their motivation without replacing this aggregate.
\end{minipage}
\medskip

\textbf{Structural holdouts.} Orthorhombic OOD accuracy is 100\%
versus 95.90\% IID (95\% material-cluster CI [94.15,97.45]); both are
100\% when restricted to matched task families. For the large-cell split,
OOD accuracy is 99.96\% [99.90,100.00] versus 96.00\% IID
[94.26,97.56]. Higher aggregate OOD scores do not establish improved
performance under shift because the evaluated family mixtures differ.
The training-scale curves appear once, in Figure~\ref{fig:scaling-story}.

\noindent\begin{minipage}{\linewidth}
\begin{figure}[H]
\centering
\includegraphics[width=\linewidth]{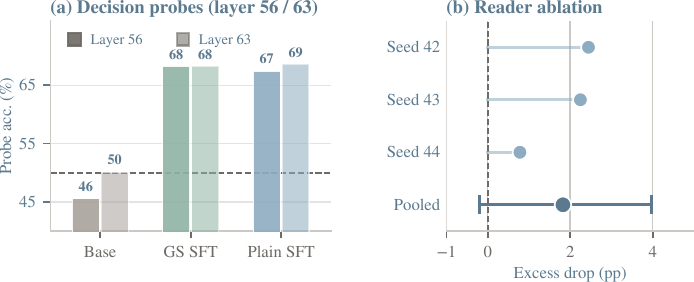}
\caption{\textbf{Probe decodability and reader-head ablation.} Probe markers distinguish layers 56 and 63; only the pooled ablation row has a stored 95\% interval.}
\label{fig:app-mechanisms}
\label{fig:probes}
\end{figure}
\smallskip
Figure~\ref{fig:app-mechanisms}(a) uses identity-isolated nested
cross-validation on 128 materials and 768 renderings. Both SFT views
improve on the base, without establishing a GraphSpace-specific advantage.
Panel (b) compares top-five reader ablation with equal-size random
controls on 341 cases; the pooled excess drop is $+1.82$~pp,
95\% CI $[-0.20,+3.98]$. Decodability and a positive point estimate
leave a candidate circuit, not unique causal localization.
\end{minipage}
\medskip

\noindent\begin{minipage}{\linewidth}
\begin{figure}[H]
\centering
\includegraphics[width=\linewidth]{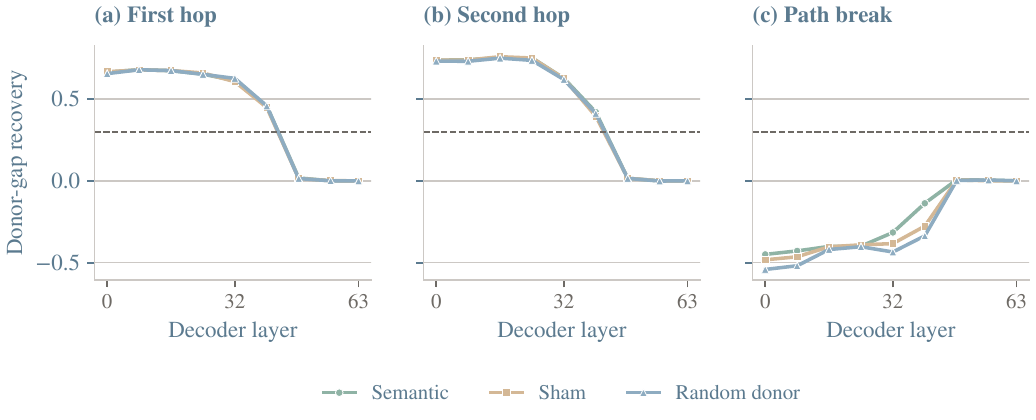}
\caption{\textbf{Activation patching across three two-hop recipient views.} The dashed 0.30 line is only the recovery-magnitude component of the joint diagnostic.}
\label{fig:app-patching}
\end{figure}
\smallskip
Figure~\ref{fig:app-patching} compares semantic donors with sham and
random controls for the GraphSpace-clean checkpoint over nine layers and
64 materials. None of the 27 layer/view cells meets the joint magnitude
and control-contrast criteria, so recovery above 0.30 does not by itself
identify a causal site.
\end{minipage}
\medskip

\subsection{Reliability, scoring, and tool references}
\noindent\begin{minipage}{\linewidth}
\begin{figure}[H]
\centering
\includegraphics[width=\linewidth]{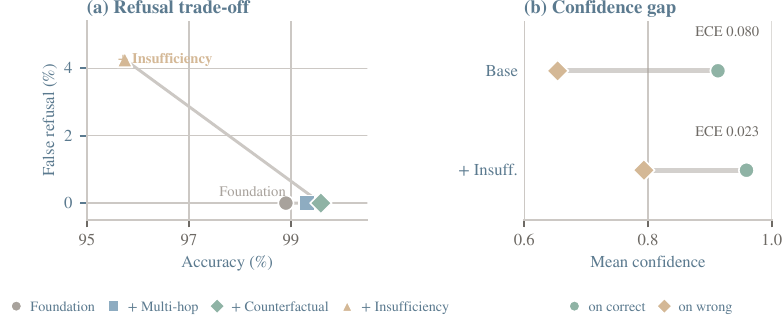}
\caption{\textbf{Refusal and confidence diagnostics.} Confidence points separate correct and incorrect answers; group-specific confidence uncertainty is unavailable.}
\label{fig:app-reliability}
\end{figure}
\smallskip
Figure~\ref{fig:app-reliability}(a) shows that insufficiency
supervision increases false refusals to 4.26\%, reducing overall accuracy
despite perfect accuracy among answered cases. Panel (b) shows lower
expected calibration error but still-high confidence on wrong answers:
better average calibration does not eliminate confident errors.
Appendix Table~\ref{tab:memorization} reports the complementary
memorization controls once; their unchanged identity, formula and
site-label contrasts do not rule out every source of contamination.
\end{minipage}
\medskip

\noindent\begin{minipage}{\linewidth}
\begin{figure}[H]
\centering
\includegraphics[width=\linewidth]{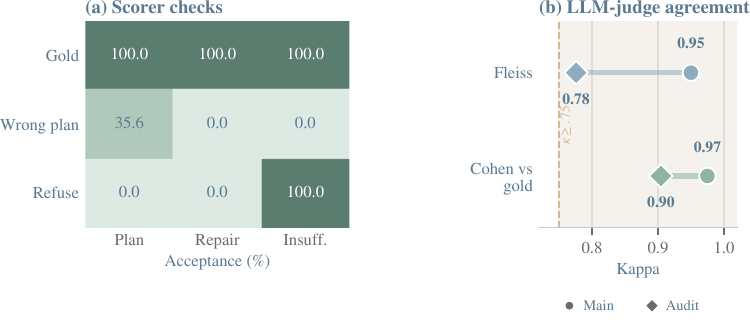}
\caption{\textbf{Scorer acceptance and LLM-judge agreement.} The judge panel separates agreement between judges from agreement with gold, and marks a 0.75 reference.}
\label{fig:app-scoring}
\end{figure}
\smallskip
Figure~\ref{fig:app-scoring}(a) verifies gold replay and null-policy
behavior. The wrong-plan policy still passes 35.6\% of plan-to-CIF cases,
a nonzero null rate that capability claims must account for. Panel (b)
shows lower judge agreement on the audit set than the main set;
these are automated, not human, raters.
\end{minipage}
\medskip

\noindent\begin{minipage}{\linewidth}
\begin{figure}[H]
\centering
\includegraphics[width=\linewidth]{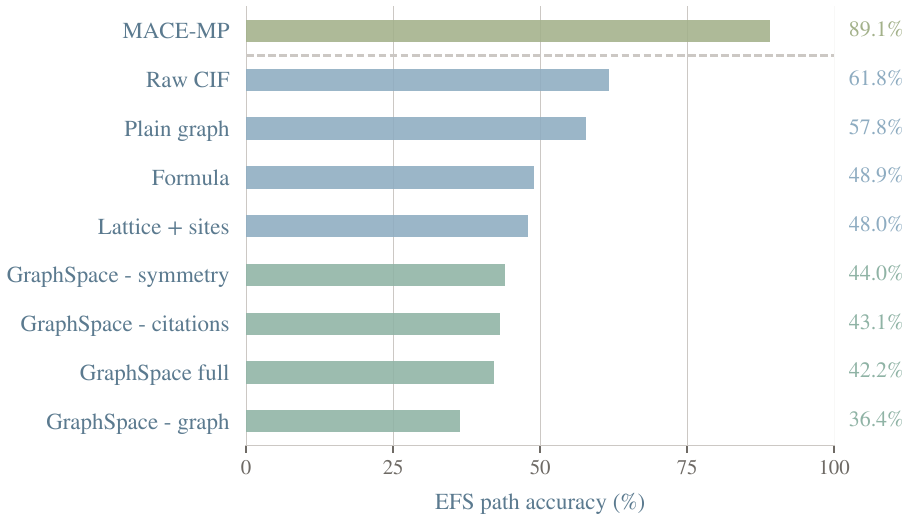}
\caption{\textbf{Professional-tool and SFT references on EFS paths.} MACE-MP uses 64 unique frame--task pairs, whereas SFT views use 225 prompts across three seeds.}
\label{fig:baselines}
\end{figure}
\smallskip
Figure~\ref{fig:baselines} keeps all eight SFT views under the same
prompt-level evaluation. MACE-MP scores $57/64$ unique frame--task
pairs, providing a professional-tool reference. Different denominators
and repeated LLM wordings prevent treating the tool--LLM gap as a
paired representation effect.
\end{minipage}
\medskip

\subsection{Failure cases and incomplete capabilities}
\noindent\begin{minipage}{\linewidth}
\begin{figure}[H]
\centering
\includegraphics[width=\linewidth]{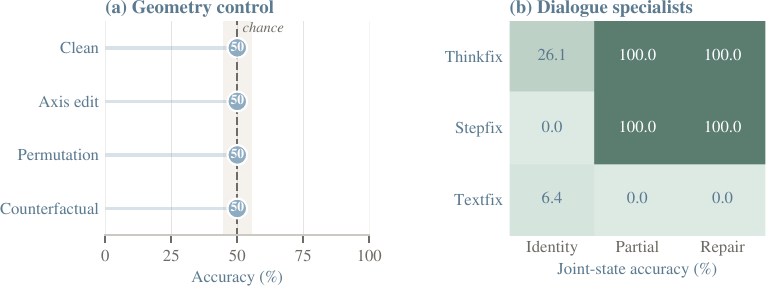}
\caption{\textbf{Geometry controls and multi-turn specialist scores.} Dialogue values are teacher-forced joint-state accuracies, not final-answer accuracies.}
\label{fig:app-capability-limits}
\end{figure}
\smallskip
Figure~\ref{fig:app-capability-limits}(a) scores 50\% in every
balanced geometry condition because all 8,192 responses select option A;
this does not establish invariance or geometry reasoning.
Panel (b) shows limited identity retrieval despite stronger
partial-structure and repair scores. Thinkfix and stepfix share
checkpoints but differ in template parity, so their scores are not a
substitute for a final-answer capability metric.
\end{minipage}
\medskip

\noindent\begin{minipage}{\linewidth}
\begin{figure}[H]
\centering
\includegraphics[width=\linewidth]{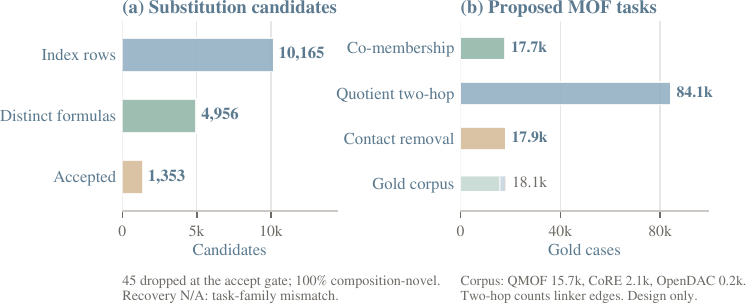}
\caption{\textbf{Recovery status and the proposed MOF task suite.} Neither panel establishes an admitted structure-generation or module-reasoning capability.}
\label{fig:app-pending}
\end{figure}
\smallskip
Figure~\ref{fig:app-pending}(a) reports compositional novelty among
1,353 accepted substitution candidates, with 45 dropped. Novelty does
not establish physical validity; recovery ran on 521 cases but is N/A
because of a training/evaluation task-family mismatch.
Panel (b) lists three candidate MOF tasks over 23,284 gold cases.
Admission has not been granted, so no hypergraph advantage is claimed.
\end{minipage}
\medskip

\clearpage
\section{Detailed experimental tables}
\label{app:detailed-tables}

These tables retain the full numerical records behind the main-text
analyses, in the order the paper uses them: the audit map and the shortcut
controls first, then the behavioural interventions on S4c, then the
representation audits. Each table is followed by a note giving its cohort,
its estimator, and what the numbers do not establish. Denominators differ
between tables by design, so rows drawn from different tables should not be
subtracted from one another.

\noindent\begin{minipage}{\linewidth}
\begin{table}[H]
\centering
\small
\caption{\textbf{CARAT audit categories and claim boundaries.}}
\label{tab:carat-gates}
\begin{tabular}{p{0.20\linewidth}p{0.34\linewidth}p{0.35\linewidth}}
\toprule
\textbf{Audit} & \textbf{Checks} & \textbf{Claim boundary} \\
\midrule
Scoring & Gold replay; null policies & Validates tested scorer contracts \\
Shortcuts & Answer masking; identity edits & Separates readback from evidence use \\
Mechanism & Probes; activation patching & No localized causal site established \\
Generalization & Element, symmetry, size, phrasing & Restricted to evaluated holdouts \\
Reliability & Refusal; confidence; judge agreement & Task- and judge-dependent evidence \\
Tool reference & MACE-MP comparison & Different evaluation denominators \\
\bottomrule
\end{tabular}
\end{table}
\smallskip
The six rows are the audit families CARAT runs, with the middle column naming
the concrete check and the right column the boundary on what a pass licenses.
Three return no binary verdict: the mechanism probes establish no localized
causal site, generalization extends only to the holdouts evaluated, and the
tool reference uses a different denominator from the LLM prompts. Each audit
targets one failure mode, and passing it establishes nothing outside its
scope.
\end{minipage}
\medskip

\noindent\begin{minipage}{\linewidth}
\begin{table}[H]
\centering
\setlength{\abovecaptionskip}{2pt}
\setlength{\belowcaptionskip}{2pt}
\caption{\textbf{Memorization controls.}}
\label{tab:memorization}
\begin{tabular}{lrr}
\toprule
\textbf{Input} & \textbf{Acc.} & \textbf{$\Delta$} \\
\midrule
Anonymous        & $0.9937$ & --- \\
Identity shown   & $0.9905$ & $-0.0032$ \\
Formula hidden   & $0.9921$ & $-0.0016$ \\
Sites relabelled & $0.9953$ & $+0.0016$ \\
Labels shuffled  & $0.4968$ & $-0.4968$ \\
\bottomrule
\end{tabular}
\end{table}
\smallskip
The controls run on 634 matched cases, with accuracy changes measured
against anonymous inputs. The tested identity and surface edits show no
detectable gain: all paired 95\% material-cluster bootstrap intervals include
zero over $10{,}000$ resamples, and the identity-visibility McNemar test gives
$p=0.727$. Shuffled labels fall to near chance, which is the positive control for the
table: it shows the metric would register an effect of this size if one were
present. These controls find no benefit from the tested cues, but do not rule
out every form of memorization.
\end{minipage}
\medskip

\phantomsection\label{app:diagnostics}
\noindent\textbf{The three anti-shortcut diagnostics.} Each failure mode
named in \S\ref{sec:eval-protocol} has its own check.
\emph{Recitation:} cases whose view prints the gold answer, such as a
space-group label, are marked \emph{recitation-eligible}; capability is
credited only on answer-absent views, and answer masking provides a direct
retest of the same questions.
\emph{Tolerance chasing:} a symmetry label depends on the tolerance used to
compute it, so the diagnostic sweeps \texttt{symprec} and checks whether
accuracy tracks the label-generation setting rather than the structure
(Figure~\ref{fig:app-tolerance}).
\emph{Pattern matching:} counterfactual edits test whether responses follow
structural changes rather than surface tokens, through an axis relabeling
that preserves the answer and neighbour edits or deliberate corruption that
change it; the response must follow the physically specified direction in
both cases. Identity visibility, formula hiding, site relabeling and label
permutation add the memorization controls of
Table~\ref{tab:memorization}.
\medskip

\noindent\begin{minipage}{\linewidth}
\begin{table}[H]
\caption{\textbf{Dominant-neighbour accuracy by cited evidence.} Citation groups are observational; the neither-cited subgroup contains only five cases.}
\label{tab:A0}
\centering
\begin{tabular}{lcc}
\toprule
\textbf{View / group} & \textbf{S4c accuracy} & \textbf{n} \\
\midrule
Grounded (all cases)                                              & 68.5\% & 499 \\
\quad cites both orbit\_link \& sibling          & 72.1\% & 165 \\
\quad cites only orbit\_link                              & 67.5\% & 329 \\
\quad cites neither                                                & 20.0\% & 5 \\
\midrule
Plain-graph (all cases)                                           & 46.3\% & 367 \\
\quad cites orbit\_link (any family)                      & n/a, rate = $0.000$ & --- \\
\bottomrule
\end{tabular}
\end{table}
\smallskip
Cases are stratified by what the model's own evidence-reference field cites.
Grounded cases citing both the orbit link and a sibling answer at $72.1\%$
against $20.0\%$ for cases citing neither, but that second group holds five
cases, so the stratification is descriptive. The plain-graph row carries a
citation rate of exactly zero because the field does not exist in that view,
which is a property of the rendering rather than of the model. Citation
groups are observational and unequally sized, so no causal effect of citing
is identified here; Table~\ref{tab:recitation} supplies that test.
\end{minipage}
\medskip

\noindent\begin{minipage}{\linewidth}
\begin{table}[H]
\caption{\textbf{Answer masking in symmetry tasks.} Each view uses 132 cases (seed 42); answer-printing views are excluded from capability claims.}
\label{tab:recitation}
\centering
\begin{tabular}{lcc}
\toprule
\textbf{View} & \textbf{Group} & \textbf{Accuracy} \\
\midrule
graphspace\_full              & Print-through   & $100.0\%$ \\
graphspace\_minus\_graph      & Print-through   & $100.0\%$ \\
graphspace\_no\_evidence\_refs& Print-through   & $100.0\%$ \\
\midrule
graphspace\_minus\_symmetry\_local & Answer-masked & $\mathbf{71.2\%}$ \\
\midrule
raw\_cif                       & Alternative     & $84.9\%$ \\
raw\_lattice\_sites            & Alternative     & $66.7\%$ \\
plain\_graph\_json             & Alternative     & $54.5\%$ \\
formula\_text                  & Alternative     & $39.4\%$ \\
\bottomrule
\end{tabular}
\end{table}
\smallskip
The three print-through views score $100\%$ because the space group appears
verbatim in their evidence. Masking the symmetry and local blocks costs
$28.8$ points and leaves GraphSpace at $71.2\%$, below raw CIF at $84.9\%$
and near lattice and sites at $66.7\%$. Because removing that block also
changes answer visibility, this is an answer-visibility diagnostic rather
than a single-token intervention at a fixed checkpoint, and the masked row
is the one a capability claim may use.
\end{minipage}
\medskip

\noindent\begin{minipage}{\linewidth}
\begin{table}[H]
\caption{\textbf{Typed-evidence masking and injection.} Deltas and 95\% bootstrap intervals use intervention-matched cases, 379 for masking and 354 for injection; baseline rows cover all cases.}
\label{tab:A1}
\centering
\begin{tabular}{lcccc}
\toprule
\textbf{Variant} & \textbf{Operation} & \textbf{S4c acc} & \textbf{$\Delta$} & \textbf{95\% CI} \\
\midrule
V0 grounded    & (baseline)                             & 68.5\% & ---     & --- \\
V1             & remove orbit-link tags                 & 66.3\% & $-2.2$  & $[-3.8, -0.6]$ \\
V2             & also remove sibling references         & 63.4\% & $-5.0$  & $[-7.4, -3.0]$ \\
\textbf{V3}    & also remove typed neighbour shell      & \textbf{25.1\%} & $\mathbf{-45.6}$ & $\mathbf{[-51.7,\,-39.6]}$ \\
\midrule
V0 plain-graph & (baseline)                             & 45.1\% & ---     & --- \\
\textbf{V4}    & inject typed edges into plain-graph & \textbf{74.3\%} & $\mathbf{+27.7}$ & $\mathbf{[+21.8,\,+33.9]}$ \\
\bottomrule
\end{tabular}
\end{table}
\smallskip
Each delta is paired on the case ids both arms scored, which is why the mask
and injection rows carry their own $n$; the full-view rows above them report
all available cases and are not a term in any of these differences. The mask leg is nested: V1 removes the orbit-link tags, V2 removes the
sibling references as well, and V3 removes the typed neighbour shell on top
of both. The first two steps cost $2.2$ and $5.0$ points against the unmasked
baseline and the third costs $45.6$, so the collapse appears only once the
shell is removed; because each row carries its own paired baseline, the step
from V2 to V3 is not itself a matched increment. The additive leg runs the other way,
injecting the same typed edges into plain graph for $+27.7$. Each
intervention is compared with its own paired baseline, and the two baselines
differ, so the full-view rows summarize all available cases and must not be
subtracted from the intervention rows below them, each of which carries its
own paired baseline.
\end{minipage}
\medskip

\noindent\begin{minipage}{\linewidth}
\begin{table}[H]
\caption{\textbf{Thinking-SFT versus matched frozen-baseline references.} Every row is scored over all 1,900 cases, non-parsing counted as errors, with material-clustered paired bootstrap intervals.}
\label{tab:A2v6}
\centering
\small
\setlength{\tabcolsep}{4.5pt}
\begin{tabular}{lcccc}
\toprule
\textbf{View} & \textbf{Baseline} & \textbf{Thinking-SFT} & \textbf{Paired $\Delta$} & \textbf{95\% CI} \\
\midrule
gs (frozen baseline)               & 34.53\% & 37.68\% & $+3.16$~pp  & $[+0.79,\,+5.46]$ \\
V1 mask orbit\_link                & 33.47\% & 37.37\% & $+3.89$~pp  & $[+1.54,\,+6.11]$ \\
V4 pg $+$ orbit\_link              & 29.42\% & 35.26\% & $+5.84$~pp  & $[+3.27,\,+8.41]$ \\
V2 mask sibling\_site\_ref         & 31.74\% & 20.58\% & $-11.16$~pp & $[-13.31,\,-9.01]$ \\
V3 mask sibling\_neighbours\_typed & 11.89\% & 4.00\%  & $-7.89$~pp  & $[-9.46,\,-6.28]$ \\
\bottomrule
\end{tabular}
\end{table}
\smallskip
The three views that still carry the evidence the question needs gain
between $3.16$ and $5.84$ points from thinking supervision, while the two
masked views lose $11.16$ and $7.89$; under those masks the supervised model
more often exhausts its token budget, as Appendix~\ref{app:paired-subsets}
shows. All five rows share one denominator of $N = 1{,}900$, so the columns
are comparable; differences between rows are still not matched
differences-in-differences, since each row masks a different field. Appendix~\ref{app:paired-subsets} gives the parse rates behind the denominator and the predicted-only figures replaced.
\end{minipage}
\medskip

\noindent\begin{minipage}{\linewidth}
\begin{table}[H]
\caption{\textbf{Reader-head ablation on 341 dominant-neighbour cases.} Teacher-forced answer scores compare the top five reader heads with equal-size random controls, under paired case bootstraps.}
\label{tab:A3}
\centering
\small
\begin{tabular}{lcccc}
\toprule
\textbf{Arm} & \textbf{heads} & \textbf{acc} & \textbf{$\Delta$ S4c drop} & \textbf{95\% CI} \\
\midrule
intact\_baseline           & 0  & 1.0000 (341/341) & $+0.00$ & --- \\
\textbf{top5\_orbit\_readers} & 5  & 0.9531 (325/341) & \textbf{$+4.69$} & $[+2.64, +7.04]$ \\
Random heads (seed 42, max) & 5 & --- & $+2.93$ & --- \\
\bottomrule
\end{tabular}
\end{table}
\smallskip
The intact baseline answers all $341$ cases correctly by construction, since
the cohort is defined as the cases the model already gets right under teacher
forcing. Ablating the five orbit-link-active reader heads costs sixteen of
them, a drop of $4.69$ points, against a maximum of $2.93$ for the best
equal-size random control in this seed. Reader-head ablation is therefore not
uniquely localized: pooled over three seeds, the excess drop relative to
random controls has an interval spanning zero
(Appendix~\ref{app:limitations}).
\end{minipage}
\medskip

\noindent\begin{minipage}{\linewidth}
\begin{table}[H]
\caption{\textbf{Canonicalization audit across eight releases.} Compression is the raw-to-canonical edge ratio on full-edge cases; dashes indicate unavailable full-edge inputs.}
\label{tab:B1}
\centering
\begin{tabular}{lccc}
\toprule
\textbf{Release} & \textbf{n} & \textbf{Compress$\times$} & \textbf{Warnings} \\
\midrule
Pilot                     & $262$     & $1.10$ & $0$ \\
Final canonical           & $2{,}000$ & $1.08$ & $0$ \\
Full (v1)                 & $2{,}000$ & ---    & $0$ \\
Full (v2)                 & $2{,}000$ & $1.36$ & $0$ \\
Hardened v2               & $2{,}000$ & $1.05$ & $0$ \\
Hardened v3 (NL rewrite)  & $2{,}000$ & $1.05$ & $0$ \\
Hardened v4               & $2{,}000$ & $1.08$ & $0$ \\
Pure-geometry (S4d)       & $2{,}000$ & ---    & $0$ \\
\midrule
\textbf{Total}            & \textbf{$14{,}262$} & --- & \textbf{$0$} \\
\bottomrule
\end{tabular}
\end{table}
\smallskip
Eight releases are sampled, $14{,}262$ cases in total, and every sampled case
satisfies the checked canonicalization conditions with zero warnings
throughout. Compression is the raw-to-canonical edge ratio and runs from
$1.05$ to $1.36$ where full edge lists are present. Two releases carry no
full-edge inputs, so the ratio is undefined for them rather than zero. The
audit covers the releases the paper uses, hardened v4 among them.
\end{minipage}
\medskip

\noindent\begin{minipage}{\linewidth}
\begin{table}[H]
\caption{\textbf{Hierarchical canonicalization audit across all releases.}}
\label{tab:B6}
\centering
\begin{tabular}{lcc}
\toprule
\textbf{Axiom} & \textbf{Result} & \textbf{Sample} \\
\midrule
A1$'$ hierarchical idempotency              & \textbf{PASS 100\%}         & 170{,}290 rows \\
A2$'$ permutation invariance (sampled)      & \textbf{PASS 100\%}         & 120 sampled \\
A3$'$ information conservation              & \textbf{PASS 100\%}         & 170{,}290 rows \\
\bottomrule
\end{tabular}
\end{table}
\smallskip
The three extended axioms are checked on the hierarchical form. Idempotency
and information conservation run over all $170{,}290$ rows of the audit
corpus; permutation invariance under module reordering is checked on $120$
sampled cases. Expansion recovers the canonical form on the audit corpus,
which is a property of the representation and alone certifies no downstream
reasoning.
\end{minipage}
\medskip

\noindent\begin{minipage}{\linewidth}
\begin{table}[H]
\caption{\textbf{Cross-material canonicalization under a physical-equivalence rubric.}}
\label{tab:B8}
\centering
\begin{tabular}{lcc}
\toprule
\textbf{Class} & \textbf{Pass/Total} & \textbf{Rate} \\
\midrule
A elemental substitution        & 33/40                   & 82.5\%          \\
C symmetry variant              & \textbf{30/30}          & \textbf{100\%}  \\
B$'$ synthetic supercell        & \textbf{30/30}          & \textbf{100\%}  \\
\bottomrule
\end{tabular}
\end{table}
\smallskip
Three rubric classes are audited. The symmetry variants and the synthetic supercells pass at $100\%$, and
elemental substitution passes 33 of 40 under the physical-equivalence
rubric. Synthetic positive controls
test nontrivial transformations; natural supercells are no substitute.
\end{minipage}
\medskip

\clearpage

\section{Analysis experiments: full protocol}
\label{app:mechanism-details}

One concern motivates the mechanistic layer specifically: a bespoke
``orbit-link reader head'' would erase the effect on ablation, putting the
advantage in one narrow circuit rather than the representation.

This appendix backs up Section~\ref{sec:mechanism} with the full
per-layer protocol. The five layers appear in the same order as the
main-text summary: correlational evidence
(\S\ref{app:A0}), causal reveal by masking and additive injection
(\S\ref{app:A1}), generative check via thinking supervision
(\S\ref{app:A2}), direct thinking-evidence alignment
(\S\ref{app:A2-gates}), and mechanistic circuit tracing
(\S\ref{app:A3}). The ASCII layout of the candidate reader circuit
(\S\ref{app:A3-diagram}) is a supplementary presentation of the
tracing results.

\subsection{Correlational evidence: the model cites what it reads}
\label{app:A0}
\label{sec:A0}

The correlational check is descriptive; the preregistered
question is whether, holding the checkpoint fixed, the model's own
\texttt{evidence\_refs} citations stratify accuracy on the hardest family
(\texttt{S4c}) with an effect exceeding a null of zero. We report the raw
stratification without a further test.

The model is fine-tuned to emit an \texttt{evidence\_refs} field in its answer
JSON that lists which typed-edge blocks it consulted (e.g.,
\texttt{orbit\_link::site::0->site::1}, \texttt{sibling\_site\_ref::site::1},
\texttt{sibling\_neighbours\_typed::site::1}). Because the citation is
SFT-learned rather than prompt-elicited free-form reasoning, and is
mechanically checkable against the input, it is a stronger correlational
handle than a written thinking trace.

Appendix Table~\ref{tab:A0} gives the stratification. In the grounded view, cases that
cite both \texttt{orbit\_link} and a \texttt{sibling} literal answer S4c at
$72.1\%$, while cases that cite neither answer at $20.0\%$: a raw $\Delta = +52.1$
percentage points. In the plain-graph view, the model cites
\texttt{orbit\_link} at rate $0.000$ across \emph{every} family (S1, S3, S4a,
S4b, S4c), because the field does not exist in that view. The citation is
real in the grounded view, and its absence in plain-graph is structural: an
information-theoretic gap in the view, not a model deficit. Whether the cited fields are causally required is a separate question, which
the correlational stratification does not address.

\subsection{Causal reveal by masking and additive injection}
\label{app:A1}
\label{sec:A1}

Before running the reveal attack we locked the
mask rubric: \textbf{strong} if any of V1/V2/V3 pushes grounded S4c into
$[40\%, 55\%]$ with a paired-bootstrap CI excluding zero;
\textbf{medium} if any mask drops S4c by $\ge 20$ pp with CI excluding zero.
The additive dual (V4) is preregistered separately as
\textbf{strong} if the injection raises plain-graph S4c by $\ge 15$ pp with
CI excluding zero.

We construct four surgical variants of the grounded view whose per-key JSON
bytes are byte identical to the baseline outside of the targeted deletion or
injection, verified by a byte-diff invariant at build time. Each intervention
uses its common cases with the baseline, with 2,000 paired-bootstrap
resamples (seed 20260819); baseline rows include all available cases.

Appendix Table~\ref{tab:A1} shows two effects. On the mask leg, deleting
\texttt{orbit\_link} alone is nearly harmless ($\Delta = -2.2$ pp), and
deleting the sibling reference on top of it costs another $\approx 3$ pp;
but deleting the typed neighbour shell (\texttt{sibling\_neighbours\_typed}),
the block that carries the neighbours' \emph{element} labels, crashes S4c
to $25.1\%$, a $-45.6$ pp drop from its paired $70.7\%$ baseline. The causal weight sits in the
element-labelled typed shell, not in \texttt{orbit\_link} per se. Under the
preregistered rubric this is a \textbf{medium} mask effect: the CI on V3
excludes zero and the effect exceeds the medium-tier $\ge 20$ pp bar. On
the additive leg, injecting the same typed edges into the plain-graph view
raises paired S4c from $46.6\%$ to $74.3\%$, $\Delta = +27.7$ pp with CI
$[+21.8, +33.9]$ excluding zero: a \textbf{strong} additive verdict. The
two legs agree in sign and are comparable in magnitude, and the additive
leg attributes the gap to representation content rather than to model
capability: the same model, given the same typed edges, exceeds the
grounded baseline.

\subsection{Generative check via thinking supervision}
\label{app:A2}
\label{sec:A2}

The generative layer asks whether \emph{training} the model to emit an explicit
oracle-chain thinking trace before the answer sharpens the same causal
signature that the reveal attack elicits from the frozen baseline. The
supervision is a $5{,}611$-row pack of \textbf{deterministic
oracle-chain thinking traces} generated directly from GraphSpace
evidence (no LLM in the loop) at the full accepted ceiling of the S4a
and S4c families, filtered through anti-shortcut, anti-readback,
oracle-chain and minimum-chain-length gates. The pack is material disjoint from the evaluation subsample by construction.

Figure~\ref{fig:thinking-evidence} traces one held-out case for material
mp-1226058. The supplied orbit link connects anchor site 2 to sibling
site 1; all twelve typed-neighbour entries are Co. With this evidence,
the model cites Co and returns the correct element. After masking the
typed shell, it instead invents O neighbours and returns O. The original
thinking also deduplicates periodic neighbours by site reference, so this
example illustrates evidence-sensitive answers, not a verified counting
procedure or proof of faithful internal reasoning. Aggregate contrasts
and alignment checks are retained in Appendix Figure~\ref{fig:thinking-stats}.

Full-parameter SFT is run on top of ckpt-\texttt{14182} (seed $42$,
LR $5\text{e-}6$, adam, linear decay, GBS $64$, $3$ epochs, TP $=8$,
$263$ optimizer steps) and the resulting checkpoint is evaluated with
the frozen \texttt{evaluate\_fixed\_predictions} scorer on the same
$1{,}900$-case gs subsample used by the correlational and reveal-attack
analyses, together with the four mask variants V1--V4.

Appendix Table~\ref{tab:A2v6} reports the paired bootstrap
(SEED $20260820$, BOOT $=2000$, paired on \texttt{eval\_id}) of the
trained checkpoint against the matched frozen-baseline references. On the intended-view (gs) eval the trained checkpoint IO improves by
$+3.16$~pp with $95\%$~CI $[+0.79, +5.46]$, scoring every non-parsing output
as an error over the common denominator of $1{,}900$; the CI excludes $0$.
Appendix~\ref{app:paired-subsets} gives this accounting alongside the
predicted-only figures that it replaces, together with the per-row parse
counts behind both.

The per-view contrasts change sign when the typed evidence is removed.
Relative to its matched frozen baseline, thinking-SFT improves IO on
unmasked GraphSpace by $+3.16$~pp, but has changes of $-11.16$~pp when
sibling references are masked and $-7.89$~pp when the typed neighbour shell
is masked. Their intervals exclude zero (Appendix Table~\ref{tab:A2v6}).
This pattern is consistent with the trained model's gains depending on the
graph evidence emphasized during supervision. All five rows now share the
denominator $N = 1{,}900$, so differences between them are computable; we report each row against its own matched frozen baseline, which is the
contrast each interval was computed for. The fixed-checkpoint causal mask-and-inject evidence remains a separate
experiment, reported in \hyperref[sec:A1]{the mask-and-inject analysis}
earlier in this appendix.

Appendix Figure~\ref{fig:thinking-stats}(a) summarizes five selected
graph-evidence replay conditions, not mask effects relative to unmasked
GraphSpace. The separate plain-graph condition has a paired IO change of
$-15.19$~pp (95\% CI $[-17.27,-13.10]$, $n=1{,}488$), so the gain does
not reach every evidence view.

\subsection{Direct thinking-evidence causal alignment}
\label{app:A2-gates}

A remaining
concern for any explicit-thinking supervision is that the emitted
\texttt{<THINKING>} block might be a post-hoc rationalization rather than the
model's real reasoning path. We tested the observable part of this with three preregistered alignment
gates (BOOT $=2000$, SEED $=20260825$; identity-clustered on
\texttt{eval\_id}), operating on the trained checkpoint's held-out gs
predictions with no additional inference or training. All three passed the
strong tier.

Appendix Figure~\ref{fig:thinking-stats}(b) displays these checks with their
individual thresholds; citation faithfulness, Jaccard overlap, and
citation--crash correlation are not treated as a common score.

(1) \textbf{Citation Faithfulness.} We regex-extract every evidence
reference the thinking cites and check it against the JSON evidence block
of the prompt. Aggregating over the $1{,}900$ held-out cases, cited
\texttt{site\_ref} tokens verify at $\mathbf{99.60\%}$, cited
\texttt{orbit\_link} anchor--sibling pairs at $\mathbf{100.00\%}$, and cited
\texttt{sibling\_neighbours\_typed} tuples at $\mathbf{92.58\%}$. Cases whose
thinking is fully faithful answer at $46.95\%$; cases with any fabricated
reference answer at $25.00\%$, a $-21.95$~pp gap that treats citation
faithfulness itself as an internal correctness signal.

(2) \textbf{Cross-view Divergence.} For every case, the thinking under gs
should diverge from the thinking under pg, because pg lacks GraphSpace-specific
fields (\texttt{orbit\_link}, \texttt{sibling\_neighbours\_typed}). Token-level
Jaccard similarity between the two thinkings has mean
$\mathbf{0.202}$ with $95\%$ CI $[0.194, 0.209]$
on $1{,}900$ paired cases, cleanly below the pre-registered $0.30$
divergence threshold: the model reshapes its stated reasoning in
response to what the view contains.

(3) \textbf{Ablation Alignment.} For each case we count the number of
\texttt{sibling\_neighbours\_typed} entries the thinking cites and, using
the same case's V3-mask prediction, record whether V3 causes an answer flip
(gs-correct $\to$ V3-wrong). Pearson correlation between citation count and
V3-induced crash is $r = \mathbf{+0.295}$ with $95\%$ CI $[+0.246, +0.342]$
(bootstrap-resampled by \texttt{eval\_id}), strong-tier above the
pre-registered $+0.20$ threshold with the CI cleanly excluding zero: the
more the model cites, the more it crashes when that specific field is
masked. The same pattern holds directionally for V1 and V2 (masked
\texttt{orbit\_link}, masked \texttt{sibling\_site\_ref}).

The three checks support a relationship between stated reasoning and
available evidence, but do not exclude post-hoc rationalization or verify
every generated step. Together with the paired IO gain, they show that
thinking-SFT benefits the intended view, that its per-view gains reverse
when key fields are absent, and that cited evidence predicts vulnerability
to masking. They do not establish an increased causal crash depth across
the distinct paired cohorts.

Behavioural evidence at the input and thinking layers is nonetheless
compatible with a single narrow reader head that could be isolated,
which the four preceding layers cannot rule out. We therefore trace
the mechanism at the weights level.

\subsection{Mechanistic circuit tracing}
\label{app:A3}
\label{sec:A3}

On the 341-case cohort where the baseline emits a correct \texttt{S4c}
answer, cites \texttt{orbit\_link}, and the parallel plain-graph prompt
lacks \texttt{orbit\_link}, we run three complementary analyses:
(1) per-head mean attention to the \texttt{orbit\_link}
key/body/sibling spans; (2) residual-stream cosine trajectories
across all 64 decoder layers plus a top-K-attended-position metric per
head; and (3) causal ablation of the pre-registered top-5
\texttt{orbit\_link}-active readers against matched random-head controls,
repeated over three seeds.

\textbf{A candidate reader circuit consistent with layers 39--47.}
A3.1 and A3.3 independently identify the same set of five heads that
attend to the \texttt{orbit\_link} body across cases: L39H15, L43H22,
L43H3, L43H23, and L47H17. A3.3's top-K-attended-position metric
corroborates this pattern: L39H15 places an \texttt{orbit\_link} token
in its top-8 attended positions in $99\%$ of cases (338/341), and the
remaining four heads do so in $75$--$91\%$ of cases. The five candidate
reader heads cluster in a compact band of the network (layers 39
through 47 of the 64-layer decoder), consistent with an
\texttt{orbit\_link} readout initiated in the mid-to-late decoder.

\textbf{Targeted reader-head ablation.} Ablating the top-5
\texttt{orbit\_link}-active heads
$\{$L39H15, L43H22, L43H3, L43H23, L47H17$\}$ drops S4c accuracy by
$+4.69$ pp with paired CI $[+2.64, +7.04]$ excluding zero, while a
random 5-head control drops accuracy by at most $+2.93$ pp. The contrast
points to a \textbf{candidate multi-head circuit} carrying the
S4c \texttt{orbit\_link} operation across a few heads and
MLP paths. Head-level concentration says where the operation
is read, not what is computed; these traces cannot separate relational
composition from direct extraction over the neighbour list.

\textbf{Paired-CI note.} The seed-42 paired treatment$-$random
contrast on the $341$-case cohort has $\Delta = +2.44$~pp with
paired-bootstrap $95\%$ CI $[+0.20, +4.74]$ (2000 iters,
material-clustered, SEED $20260820$), excluding zero at $\alpha = 0.05$
with a $+0.20$~pp margin. A three-seed replication (seeds $42/43/44$)
has since completed and honestly \emph{widens} the interval: the pooled
three-seed point estimate is $\Delta = +1.82$~pp with CI $[-0.20, +3.98]$,
which does not exclude zero. We therefore leave the mechanism unlocalised, recorded as DISTRIBUTED in the
verdict log: the per-head
importance concentration in Appendix Table~\ref{tab:A3} identifies a compact
candidate reader set, and the circuit-level causal margin locates that set
without yet resolving a single causal site. The candidate reader circuit is
reported as a representation-level pointer rather than a certified circuit.

\subsection{ASCII spatial layout of the reader circuit}
\label{app:A3-diagram}

The ASCII flow below re-plots \hyperref[sec:A3]{the circuit-tracing analysis}'s localization of the
\texttt{orbit\_link} reader circuit and prints, at each stage, the
numbers that stage produced.

\begingroup
\footnotesize
\begin{verbatim}
INPUT PROMPT --- 341 S4c cases
                 orbit_link cited in prediction (100% cohort)

    |  A3.1 attention-head importance (16 full-attn layers x 24 heads)
    |     top-5 by (orbit_body - random_control) attention:
    |         L39 H15, L43 H22, L43 H3, L43 H23, L47 H17
    |
    |  A3.3 top-K attention membership (independent metric)
    |     same 5 heads consistently include an orbit_body token in top-8:
    |         L39 H15 (99%), L43 H23 (91%), L43 H22 (89%),
    |         L47 H17 (78%), L43 H3  (75%)
    |
    v
STAGE-A: 5 candidate reader heads consistent with a compact layer band
         (39-47). Two independent metrics agree on the same head set.

    |  A3.4 causal head ablation (teacher-forced at "dominant_element":")
    |     intact_baseline    acc = 1.0000 (341/341)
    |     TOP-5    ablated:  acc = 0.9531  ->  Delta_S4c drop = +4.69 pp
    |                        95% CI [+2.64, +7.04] excludes 0 -> causal
    |     Random 5-head controls (seed 42 maximum): +2.93 pp
    |
    v
STAGE-B: targeted ablation exceeds the seed-42 random-control drop,
         but this does not establish an excess effect across seeds.
         The result remains a candidate circuit (three-seed pooled
         Delta = +1.82pp CI [-0.20, +3.98] does not exclude zero:
         DISTRIBUTED).


                 ---- unified spatial mechanism picture ----
GraphSpace's advantage on S4c is consistent with a CANDIDATE MULTI-HEAD
READER CIRCUIT over EXPLICIT STRUCTURED TYPED RELATIONS
(orbit_link -> sibling_site_ref -> sibling_neighbours_typed). The five
identified reader heads sit in the layer-39-through-47 band; ablating
them CI-verifiably degrades S4c accuracy, and the operation appears to
be realized jointly across this specialized head set and adjacent MLP
paths -- a signature consistent with learned symbolic-relational
computation over the
GraphSpace representation.
\end{verbatim}
\endgroup

\clearpage

\section{Formal properties of GraphSpace: full audits}
\label{app:formal-details}

This appendix backs up Section~\ref{sec:formalization} with the full
per-audit protocol. The seven audits follow the main-text summary in
order: idempotency and information-conservation (\S\ref{app:B1B2}),
manifest pin and redundancy classification (\S\ref{app:B3B4}),
canonical training ablation (\S\ref{app:B5}), hierarchical canonical
form (\S\ref{app:B6}), injectivity argument (\S\ref{app:B7}),
cross-material canonical audit (\S\ref{app:B8}), and edge-case
coverage (\S\ref{app:B9}). \S\ref{app:B7-counterexamples} collects the
four counterexamples that mark the boundary of the injectivity
argument.

\subsection{Canonical form: idempotency and information-conservation audit}
\label{app:B1B2}\label{sec:B1B2}
A canonical form is admissible only if
it satisfies A1 (idempotency), A2 (permutation invariance), A3
(information conservation over the element multiset and the
$(\text{distance}, \text{element pair})$ multiset), and A6 (unambiguous
sort key), on \emph{every} sampled case of \emph{every} current release,
with zero warnings.

We define a canonical form by a three-key site order
$(\text{Wyckoff letter},\, \text{periodic number},\, \text{frac-coord
lexicographic})$, an edge grouping by
$(\text{src}, \text{dst}, \text{round}_4(d), \{e_i, e_j\})$ with an
explicit \texttt{multiplicity} field, and a periodic-boundary clamp
$\texttt{\_wrap\_frac\_to\_unit}(5)$ that folds
$v = 1.0$ back to $0.0$ (to keep the frac-coord sort key
idempotent at the boundary). \texttt{cutoff} is pinned at $6.0\,\text{\AA}$
and \texttt{symprec} at $10^{-3}$ in every manifest.

Appendix Table~\ref{tab:B1} gives the result: $14{,}262 / 14{,}262$ cases
pass idempotency, $14{,}262 / 14{,}262$ pass information conservation,
zero warnings. The most extreme single-case compression is
$52 \to 6$ canonical rows (an $8.67\times$ reduction) on a
Cs--Cl case where eight periodic-image
duplicates of the same $3.5885\,\text{\AA}$ bond were printed
verbatim.

Figure~\ref{fig:canonical-example} shows a separate, fully paired Rb
record with stored coordinates and lattice; the drawn graphs are schematic,
not reconstructions of its atomic geometry. Its 22 edge instances become
16 weighted rows, whose multiplicities sum to 22. Recomputing the
canonical form reproduces the archived output exactly and a second
application leaves it unchanged. This is a concrete representation audit,
not a claim about the model's downstream reasoning.

\subsection{Manifest pin and redundancy classification}
\label{app:B3B4}\label{sec:B3B4}
We pin \texttt{cutoff\_angstroms}, \texttt{symprec}, and
\texttt{canonical\_order\_v1} in every current release's manifest with
byte-preserving edits (old manifest checksum retained; new checksum
recorded separately). We then classify all
\texttt{MAT\_*} evidence blocks into two kinds. \textbf{Information
redundancy} means a block is exactly recomputable from a lower-level
block (\texttt{MAT\_GRAPH} from \texttt{MAT\_STRUCTURE} $+$ cutoff;
\texttt{MAT\_WYCKOFF} from \texttt{MAT\_STRUCTURE} $+$ \texttt{symprec};
\texttt{orbit\_link} from \texttt{MAT\_WYCKOFF}). This kind is
\emph{intentional and useful}: it is the reason a token-based model can
answer without doing message-passing at forward time.
\textbf{Field duplication} means the same edge appears multiple times
verbatim (as in the Cs--Cl edge-list example), which is a genuine, compressible
redundancy that our multiplicity field collects into a single
canonical row. Aggregate pooled edge compression across releases is
$1.05$--$1.36\times$ ($\approx 10\%$ byte drop), concentrated in the
S1 family (the $8.1\%$ of cases carrying a full edges list),
with a top single-case reduction of $52 \to 6$ rows ($8.67\times$).

\subsection{Canonical training ablation}
\label{app:B5}\label{sec:B5}
The training-time ablation that isolates the effect of canonicalization
on downstream IO trains two matched arms end-to-end on the same
underlying training rows: arm A on the expanded evidence form and arm
B on the canonical form. Both arms complete real full-parameter SFT
(seed $42$, GBS $64$, LR $5\text{e-}6$, TP $=8$) and are evaluated on
the frozen $77{,}514$-case \texttt{crystal\_reasoning\_eval\_final} eval
with the fixed scorer; the paired bootstrap ($n = 76{,}446$ paired
\texttt{eval\_id}s, BOOT $=2000$, SEED $=20260819$) yields

\[
\Delta_{\mathrm{IO}}\;(\text{canonical}-\text{expanded})
\;=\;+11.77\text{~pp},\quad
95\%~\text{CI}\;=\;[+11.46,\;+12.06]
\]

with the CI excluding zero. Canonicalization therefore yields a
paired IO improvement at training time on the same underlying rows
and the same downstream evaluation, carrying the audit-level results
of the \hyperref[sec:B1B2]{canonical-form audit} and
\hyperref[sec:B3B4]{redundancy classification} through to model
quality.

\subsection{Hierarchical canonical form}
\label{app:B6}\label{sec:B6}
Extended axioms A1$'$ (hierarchical
idempotency), A2$'$ (permutation invariance under module reordering),
and A3$'$ (information conservation via \texttt{expand\_hierarchical}) are
all required to pass at $100\%$ on every row of the release corpus.

The hierarchical form groups sites by Wyckoff orbit, printing one full
representative site plus \texttt{orbit\_link} pointers to the rest of
the orbit. This is a forward-extensible refinement of the canonical
form for large unit cells, and it preserves the same lemmas as the
flat canonical form (\S\ref{app:B7}). Appendix Table~\ref{tab:B6}
reports the audit over all $170{,}290$ rows of the current release
corpus.

Every hierarchical-canonicalization pass round-trips: repeated
application is a fixed point (A1$'$), reordering the input's modules
yields byte-identical output (A2$'$), and \texttt{expand\_hierarchical}
recovers the flat canonical evidence with no loss (A3$'$). The
hierarchical form is a proper refinement of the flat canonical
form: it preserves injectivity and groups the representation by
Wyckoff orbit.

The audits above observe no collision on $14{,}262$ base and
$170{,}290$ hierarchical rows, but byte-checks cannot in principle rule
out collisions on structures the corpus has not sampled. We therefore
also set out the constructive argument, and state where it stops.

\subsection{Injectivity argument and its boundaries}
\label{app:B7}\label{sec:B7}
Let a legal
\emph{Structure} be a triple $(L, S, Z)$: a lattice basis
$L \in \mathbb{R}^{3\times3}$ with $\det L > 0$, a finite site list
$S = \{(f_k, e_k)\}$ of fractional coordinates $f_k \in [0,1)^3$ and
element species $e_k$, and a space-group assignment $Z$ obtained from
\texttt{spglib} at $\texttt{symprec} = 10^{-3}$. Let a legal
\emph{CGE} object be the canonical graph evidence emitted by
\texttt{src/materials\_llm/graphspace/canonical.py} under
\texttt{canonical\_version} v1: sites ordered by the three-key
\texttt{(wyckoff\_letter, periodic\_number, frac\_coord\_lex)}, and
edges grouped by $(\mathrm{src}, \mathrm{dst},
\mathrm{round}_4(d), \{e_i, e_j\})$ with an explicit
\texttt{multiplicity} field, at a $6.0$~\AA{} neighbour cutoff. Two
Structures are equivalent when they are related by the gauge group
$G = G_{\text{trans}} \times G_{\text{perm}} \times
G_{\text{wyckoff}} \times G_{\text{lattice-basis}}$: origin shift,
site relabelling, Wyckoff representative, and lattice basis. The map
of interest is $T: \mathrm{Structure} / G \to \mathrm{CGE}$.

Injectivity of $T$ rests on four lemmas: canonical site order is
unique per equivalence class (L1); the Wyckoff-orbit partition
determines the space group up to conjugation (L2); the compressed edge
multiplicities preserve the raw pairwise distance multiset (L3); the
element species per site is preserved (L4). L4 is immediate, since
species are copied verbatim into the canonical row. L2 is standard
crystallography once $Z$ is fixed. L1 and L3 are the steps that hold conditionally, and their two conditions arise for different reasons, one from the
coordinate convention, one from quantization.

\textbf{Where the argument is conditional, stated precisely.} Three steps
hold only under conditions the informal version leaves implicit, so we state
each condition here.

First, wrapping coordinates into $[0,1)^3$ does not make the
three-key order translation invariant. An origin shift that keeps
every coordinate inside $[0,1)$ changes no coordinate's wrap status
and still permutes the lexicographic order: $(0.1, 0.2)$ and
$(0.3, 0.4)$ sort the same way, but a shift of $0.75$ sends them to
$(0.85, 0.95)$ and $(0.05, 0.15)$ and reverses them. Wrapping gives a
canonical representative of each coordinate, not a canonical origin.
Translation invariance in the implementation comes entirely from
\texttt{spglib} standardization, which fixes the origin along with the
cell before the sort runs. The order is therefore invariant relative to
that standardization and to its $10^{-3}$ tolerance, and $T$ is defined
on standardized cells rather than on arbitrary presentations. We have
not proved that the standardization itself is unique at that tolerance.

Second, $\mathrm{round}_4$ is lossy and no separation condition
recovers what it discards. A distance separation larger than
$10^{-4}$~\AA{} prevents two distinct edges from collapsing into one
canonical row, which is a statement about collisions; it says nothing
about recovering the pre-rounding value, which is simply gone. The
canonical form is therefore lossless with respect to a
$10^{-4}$~\AA{} quantization of the geometry, not with respect to the
geometry. Any reconstruction from it inherits that quantization.

Third, the round-trip audits do not test structure recovery. The A3
audit checks that the canonical evidence preserves the element multiset
and the distance multiset it was built from; \texttt{expand\_hierarchical}
checks that the hierarchical form expands back to the flat canonical
form byte-for-byte. Both are serialization round trips between
representations we define. Neither recovers a crystal from compressed
distances, and we should not have linked them to that claim.

\textbf{What we therefore claim.} Three separate properties are worth
distinguishing, and only the first two are established here.
GraphSpace is a deterministic serialization: the same standardized
input always produces the same bytes, idempotently and independently of
input ordering, verified at $14{,}262/14{,}262$ and
$170{,}290/170{,}290$ rows. It is lossless for the fields it stores, at
a $10^{-4}$~\AA{} quantization, verified by the same round trips.
It is \emph{not} shown to determine the crystal: we give no
reconstruction of a structure from a canonical record and no argument
that two inequivalent structures cannot share one. Injectivity holds on
the audited corpus as an observed fact, with no collision in
$184{,}552$ rows, and that is the whole of the evidence. No claim made in
Section~\ref{sec:experiments} or~\ref{sec:mechanism} depends on anything
stronger than that observation.

Each lemma is paired with a counterexample that names the field whose
removal would break it. CE-2 removes element species and the
element-tagged typed neighbour list, which breaks L4 and hence
injectivity for any task that depends on element identity. That such
tasks become unidentifiable from the representation does not entail
that a trained model's accuracy must fall: a model could still answer
from priors or corpus statistics. The V3 crash of $-45.6$~pp on
\texttt{S4c} and the V4 recovery of $+27.7$~pp
(\hyperref[sec:A1]{the mask-and-inject analysis}) are measurements
that happen to agree in sign with the identifiability boundary; they
are not predictions derived from it. CE-3 is witnessed by the audit's
most extreme case ($52 \to 6$ canonical rows), CE-4 by the
periodic-boundary clamp (\texttt{\_wrap\_frac\_to\_unit}) that gives
post-clamp idempotency at $14{,}262/14{,}262$, and CE-1 by the
hardened v4 release, which withholds Wyckoff labels by design for
anti-readback purposes and so narrows $T$'s domain to an equivalence
class that additionally quotients by space-group choice.

\subsection{Cross-material canonical audit}
\label{app:B8}\label{sec:B8}
Under a \textbf{physical} standard 
skeleton (site count, frac multiset, connectivity) preserved, distances
allowed to shift by species-dependent radii, a canonical form is
cross-material consistent if pairs related by an interpretable
transformation (elemental substitution, supercell scaling, symmetry
variation) canonicalize to a structurally consistent output. We audit
three positive rubrics: Class A elemental substitution ($40$ pairs),
Class C symmetry variant ($30$ pairs), and Class B$'$ a synthetic
$2 \times 1 \times 1$ supercell positive-control tiling of real
crystals ($30$ pairs) that isolates the canonical scaling axiom.

Appendix Table~\ref{tab:B8} summarizes the audit. The two positive-control
classes (symmetry variants Class C and synthetic supercell tilings
Class B$'$) both pass at $100\%$, showing that the canonical spec
is cross-material consistent on symmetry-related and supercell-related
transformations. Elemental substitution (Class
A) passes at $82.5\%$; the skeleton (site count, frac multiset,
canonical order pattern) is preserved on all $40$ pairs, with the
seven residual anomalies attributable to species-dependent bond-length
shifts that exceed the physical tolerance on same-frac different-cell
pairs.

\textbf{Class B corpus note.} The natural-supercell Class B row
currently draws from a small pilot corpus ($30$ cases); we report its
aggregate pass rate for completeness but rely on the size-matched
positive control B$'$-$30$ ($30/30$ pass) for the certified consistency
claim. Enlarging Class B to $N \ge 100$ natural supercells is a
scheduled corpus expansion.

\subsection{Edge-case coverage}
\label{app:B9}\label{sec:B9}
We further specify canonical forms for four schema-level edge cases:
(1) supercells (\texttt{supercell\_factor} $+$ primitive-cell canonical);
(2) point defects (vacancy / interstitial / substitution 10-tuple total
order); (3) partial occupancy disorder (\texttt{species\_distribution}
normalized plus 12-char fingerprint tie-break, fail-closed on occupancy
sum $\ne 1$); (4) multicomponent solid solutions
(\texttt{orbit\_id}-prefixed tiebreakers). The v2 wrapper is an
\emph{additive extension}: 25 v2-specific tests plus 22 legacy tests (9
v1 spec, 13 hierarchical) all pass ($47/47$ green). On the current 8
releases, v2 output is byte identical to v1 output on $3{,}762 / 3{,}762$
sampled rows, and the current releases contain zero edge-case rows;
v2 is thus a forward-compatible extension, inert on the current releases and
active as soon as a future one includes edge-case schemas.

\subsection{Boundary counterexamples table}
\label{app:B7-counterexamples}

Table~\ref{tab:app-B7} lists the four counterexamples from
\hyperref[sec:B7]{the injectivity argument}, the lemma each depends
on, and the release or experiment that witnesses it.

\begin{table}[H]
\caption{\textbf{Boundary counterexamples, affected lemmas, and empirical witnesses.}}
\label{tab:app-B7}
\label{tab:B7}
\begin{center}
\begin{tabular}{cllc}
\toprule
\textbf{CE} & \textbf{Removed field}        & \textbf{Lemma}        & \textbf{Witness} \\
\midrule
CE-1 & Wyckoff labels                & L2 space group    & v4 anti-readback pass \\
CE-2 & Element, typed neighbours     & L4 element        & V3 $-45.6$; V4 $+27.7$ \\
CE-3 & Edge multiplicity             & L3 distance mset  & $52 \to 6$ rows \\
CE-4 & Boundary clamp                & L1 site order     & $14{,}262/14{,}262$ pass \\
\bottomrule
\end{tabular}
\end{center}
\end{table}

\section{Metric definitions and result accounting}
\label{app:accounting}

This appendix makes the headline numbers recomputable. It states the
scoring formulas and their denominators, gives the composition of both
evaluation sets, and lists every exclusion rule that removes an item
from a reported denominator.

\subsection{Scoring formulas}
\label{app:metric-defs}

\textbf{Resampling units.} The broader benchmark uses hard-slice stratified
bootstrap contrasts with $10{,}000$ resamples. The input-intervention
comparisons in the \hyperref[sec:A1]{mask-and-inject} and
\hyperref[sec:A2]{thinking-supervision} analyses use $2{,}000$ paired
case-bootstrap resamples. The two are not interchangeable, because they
resample different units and answer different questions.

Every model output is a JSON object whose \texttt{answer} field holds
one or more named sub-fields. Let item $i$ have $F_i$ gold answer
sub-fields of which the prediction matches $m_i$ after the frozen
scorer's per-field normalization. Three quantities are used, over a set
$\mathcal{I}$ of $N = |\mathcal{I}|$ items:
\begin{align*}
\text{answer-exact rate} &= \frac{1}{N}\sum_{i \in \mathcal{I}} \mathbf{1}[m_i = F_i], \\
\text{answer-field accuracy} &= \frac{1}{N}\sum_{i \in \mathcal{I}} \frac{m_i}{F_i}, \\
\text{macro-hard-slice accuracy} &= \frac{1}{|\mathcal{S}|}\sum_{s \in \mathcal{S}} \frac{1}{|\mathcal{I}_s|}\sum_{i \in \mathcal{I}_s} \frac{m_i}{F_i}.
\end{align*}
The first is strict: an item counts only if every sub-field matches.
The second gives partial credit within an item and weights items
equally. The third is the preregistered primary metric: it averages
the second over the $|\mathcal{S}| = 6$ hard slices with equal weight
per slice, so a slice of 72 items counts as much as a slice of 266.
The reference implementation is
\texttt{macro\_hard\_slice\_score} in
\texttt{src/materials\_llm/graphspace/p8\_paired\_statistics.py}.

Table~\ref{tab:overall} reports answer-exact rate, because that is the
metric under which the per-family cells and the row totals are
mutually consistent. Table~\ref{tab:family} and
Section~\ref{sec:experiments} report macro-hard-slice accuracy, because
that is the metric the decision rule was preregistered on.
Table~\ref{tab:metric-map} gives all three for all eight views, so any
number quoted under one regime can be located under the other.

\begin{table}[H]
\caption{\textbf{The same eight views under all three metrics} (seed 42, $n{=}1{,}024$).}
\label{tab:metric-map}
\centering
{\setlength{\aboverulesep}{0pt}\setlength{\belowrulesep}{0pt}%
\begin{tabular}{lccc}
\toprule
\rowcolor{HeaderColor}
\textbf{View} & \textbf{Answer-exact} & \textbf{Field acc.} & \textbf{Macro-hard-slice} \\
\midrule
GraphSpace full          & 0.956 & 0.978 & 0.956 \\
\quad $-$~citations      & 0.876 & 0.938 & 0.926 \\
\quad $-$~graph          & 0.805 & 0.913 & 0.899 \\
\quad $-$~symmetry       & 0.922 & 0.956 & 0.944 \\
Raw CIF                  & 0.761 & 0.903 & 0.899 \\
Plain graph              & 0.821 & 0.899 & 0.891 \\
Lattice $+$ sites        & 0.682 & 0.849 & 0.842 \\
Formula                  & 0.542 & 0.772 & 0.783 \\
\bottomrule
\end{tabular}}
\end{table}
\smallskip
The GraphSpace full row agreeing to three decimals across the strict
and macro columns ($0.9561$ and $0.9560$) is a coincidence of that
view's uniform per-slice scores, not a transcription error. The gap
between regimes is largest for the formula view ($0.542$ against
$0.783$) because partial credit and equal slice weighting both favour
a view that fails whole answers but recovers individual sub-fields.

\subsection{Composition of the 1,024-question benchmark}
\label{app:benchmark-composition}

The matched benchmark contains $1{,}024$ questions drawn from $562$
distinct Materials Project entries. The clustering unit for every
bootstrap interval in the paper is \texttt{split\_group\_id}, one
identifier per source material; a material contributes between one and
seven questions (mean $1.82$). The seven task families map onto the
six hard slices as shown in Table~\ref{tab:slice-map}. Two families
share the adversarial slice, and two share the structure-reasoning
slice; no other slice mixes families. The eight views render the same
$1{,}024$ \texttt{query\_variant\_id} values, so every cross-view
comparison in Tables~\ref{tab:overall} and~\ref{tab:family} is paired on
identical questions with no missing cells.

\begin{table}[H]
\caption{\textbf{Task families, hard slices, and item counts} in the 1,024-question benchmark.}
\label{tab:slice-map}
\centering
{\setlength{\aboverulesep}{0pt}\setlength{\belowrulesep}{0pt}%
\begin{tabular}{llr}
\toprule
\rowcolor{HeaderColor}
\textbf{Hard slice} & \textbf{Task family} & \textbf{Items} \\
\midrule
CIF generation plan   & CIF plan reconstruction     & 148 \\
EFS path              & EFS path reasoning          & 72 \\
Insufficient or adv.  & Evidence sufficiency        & 174 \\
Insufficient or adv.  & EFS path reasoning          & 24 \\
MOF                   & MOF module property         & 136 \\
Standard              & Graph construction          & 204 \\
Structure reasoning   & Local environment           & 134 \\
Structure reasoning   & Symmetry reasoning          & 132 \\
\midrule
Total                 &                             & 1{,}024 \\
\bottomrule
\end{tabular}}
\end{table}

\subsection{Composition of the readback-hardened healthy pool}
\label{app:healthy-pool}

The healthy pool is a separate and much larger set. It is released as
\texttt{crystal\_\allowbreak reasoning\_\allowbreak eval\_\allowbreak
hardened\_\allowbreak v4\_\allowbreak 20260814} and contains
$77{,}514$ questions built from $8{,}001$ held-out materials, rendered
into two views (GraphSpace and plain graph) that carry identical
question text and differ only in how the evidence is presented. The
build scanned $79{,}057$ evidence records, skipped $71{,}056$ that
belong to the training or validation partitions, read all $8{,}001$
test-partition records, and quarantined none. Its $77{,}514$ rows carry $76{,}510$ distinct \texttt{eval\_id}s, 907 of
them appearing twice, and every stage of the pipeline keys by
\texttt{eval\_id} and so keeps one row per id. Across three seeds that
gives $3 \times 76{,}510 = 229{,}530$ candidate paired items, of which
$n = 229{,}485$ pair across both views, the remaining $45$ lost to
generation-side failures; that is the denominator of the $+0.193$
contrast in Section~\ref{sec:experiments}. It is not the
$1{,}024$-question benchmark and shares no questions with it: the
benchmark measures seven applied task families under the preregistered
decision rule, while the healthy pool measures seven synthetic
structure-reasoning families in which the answer token is withheld
from both views by construction. Table~\ref{tab:healthy-families}
gives the families, their sizes, the majority-class baseline that a
model could reach by always guessing the modal answer, and the score
of the parameter-free list rule of
Appendix~\ref{app:pointer-blind}, which solves the four sibling and
two-hop families without following any pointer. The remaining three
families ask about the structure as a whole and have no pointer for
that rule to ignore, so it does not apply to them.

\begin{table}[H]
\caption{\textbf{Healthy-pool families, item counts, and majority baselines.} Release v4, test partition.}
\label{tab:healthy-families}
\centering
\small
{\setlength{\aboverulesep}{0pt}\setlength{\belowrulesep}{0pt}%
\begin{tabular}{llrrr}
\toprule
\rowcolor{HeaderColor}
\textbf{Family} & \textbf{What is asked} & \textbf{Items} & \textbf{Majority} & \textbf{List rule} \\
\midrule
S1 & Count distinct Wyckoff orbits            & 6{,}555  & 0.101 & --- \\
S2 & $k$-th neighbour, then its own CN        & 23{,}993 & 0.272 & 0.949 \\
S3 & Element with max mean coordination       & 7{,}021  & 0.069 & --- \\
S4a & Anchor CN via its sibling               & 10{,}197 & 0.316 & 0.881 \\
S4b & Anchor second shell via its sibling     & 9{,}769  & 0.224 & 0.996 \\
S4c & Anchor dominant element via its sibling & 10{,}197 & 0.505 & 0.932 \\
S4d & Match candidate on three attributes     & 9{,}782  & 0.255 & --- \\
\midrule
   & Total                                    & 77{,}514 &       &    \\
\bottomrule
\end{tabular}}
\end{table}

\textbf{What is compared, and how it is weighted.} The healthy-pool
number is a paired difference at fixed weights: for each item $i$
present in both views we take $g_i - p_i \in \{-1, 0, +1\}$, the
GraphSpace-view correctness minus the plain-graph-view correctness,
and report the unweighted mean over all paired items pooled across the
three seeds. There is no family reweighting, so the pool is
effectively weighted by family size and S2 carries $31\%$ of the mass.
The comparison baseline is the plain-graph rendering of the identical
question, evaluated by the identical frozen scorer at the identical
checkpoint; it is not a majority baseline and not a different model.
Because the two views are compared within a family, the per-family majority
baseline in Table~\ref{tab:healthy-families} cancels from a subtractive
normalisation, so $(a_1 - b) - (a_2 - b) = a_1 - a_2$ and that normalisation
leaves the raw difference unchanged. A rescaling normalisation, dividing by
$1 - b$, does not: it multiplies the difference by $1/(1-b)$ and is not what
we report. The bootstrap resamples the pooled per-item differences, so the interval
covers sampling of questions rather than of materials, and records sharing a
source material are not held together. It does not cover training randomness
either: each seed is one training run, and three runs do not support a
variance component for it. Holm correction is applied
across the per-family hypotheses within a table, not across tables, and the
corrected decisions are the ones the verdict column reports. The $229{,}485$ paired predictions are three seeds over $76{,}510$ question
ids drawn from $8{,}001$ held-out materials, far fewer independent materials
than items.

\textbf{Pairing key, bootstrap, and exclusions.} Items are paired on
\texttt{eval\_id.split('::')[1:]}, the same key used by
\texttt{score\_crystal\_hardened\_paired.py}. Intervals are paired
nonparametric bootstraps with $B = 2{,}000$ resamples at the fixed
preregistered seed $20260813$; the seed was set before the runs and
not re-drawn. A cell is scored as a GraphSpace win when the interval
lower bound exceeds zero, a plain-graph win when the upper bound falls
below zero, and a tie otherwise; the tier thresholds (strong at
$\geq 6/8$ training views, medium at $4$--$5/8$, weak below $4$) were
fixed in advance. Three seeds of the release's $76{,}510$ distinct
\texttt{eval\_id}s give $229{,}530$ possible pairs per training view, and
the realized counts in Table~\ref{tab:healthy-deltas} range from
$227{,}106$ to $229{,}530$, that is $98.94$--$100\%$, so the largest
dropout on any arm is $1.06\%$. Quoting the ceiling as
$3 \times 77{,}514 = 232{,}542$ would understate the yield, because
$1{,}004$ of those rows per seed are second copies of an id the
pipeline keeps only once. Beyond that deduplication, items drop out
only when one of the two views produced no parseable prediction for
that \texttt{eval\_id}, which is a generation-side failure and not a
content filter. The row
census shows per-run yields of about $77{,}027$ of $77{,}514$
throughout, with one outlier at $74{,}519$ (plain-graph view, seed 44,
graphspace-minus-graph arm), which is what makes that arm's total lower than
the rest.

Three properties of the release bound how much such dropout can bias
the comparison. Each family recomputes its gold answer a second time
through an independent oracle path and requires equality, so every
emitted case is solvable by construction. Every drop condition applied
at build time, whether oracle disagreement, a tie between candidates,
an ambiguous distractor, or insufficient labels, removes the case from
both views; no condition was ever applied to one view alone. The gold
token is never rendered next to its answer key in either view, and all
symmetry, Wyckoff, and crystal-system fields are withheld. Inference dropout remains view asymmetric in principle, but against the same
deduplicated ceiling of $229{,}530$ it affects at most $1.06\%$ of pairs on
any arm, whereas the smallest reported effect is $13.6$ pp.

\begin{table}[H]
\caption{\textbf{Healthy-pool paired difference by training view}, pooled over seeds 42/43/44.}
\label{tab:healthy-deltas}
\centering
{\setlength{\aboverulesep}{0pt}\setlength{\belowrulesep}{0pt}%
\begin{tabular}{lrrl}
\toprule
\rowcolor{HeaderColor}
\textbf{Training view} & \textbf{Paired $n$} & \textbf{$\Delta$} & \textbf{95\% CI} \\
\midrule
GraphSpace full           & 229{,}485 & $+0.1933$ & $[+0.1913, +0.1953]$ \\
\quad $-$~citations       & 229{,}528 & $+0.1973$ & $[+0.1952, +0.1993]$ \\
\quad $-$~graph           & 227{,}106 & $+0.1678$ & $[+0.1659, +0.1696]$ \\
\quad $-$~symmetry        & 229{,}498 & $+0.1993$ & $[+0.1974, +0.2014]$ \\
Raw CIF                   & 229{,}530 & $+0.1363$ & $[+0.1346, +0.1380]$ \\
Plain graph               & 229{,}491 & $+0.2027$ & $[+0.2007, +0.2048]$ \\
Lattice $+$ sites         & 229{,}479 & $+0.2026$ & $[+0.2007, +0.2046]$ \\
Formula text              & 229{,}454 & $+0.2111$ & $[+0.2092, +0.2130]$ \\
\bottomrule
\end{tabular}}
\end{table}
\smallskip
The row quoted in Section~\ref{sec:experiments} is the GraphSpace-full
training view, $+0.1933$ over $n = 229{,}485$. The eight rows are eight
separately trained checkpoints, not eight slices of one run, which is
why the claim is a win on $8/8$ training views rather than on $8$ cells
of one matrix. Resolving each row into its seven families gives the $56$
cells whose seed agreement is $51/56$; two of them favour the
plain-graph view, S1 under the formula-trained arm and S2 under the
GraphSpace-trained arm.

\subsection{Pointer-blind solvability audit}
\label{app:pointer-blind}

Section~\ref{sec:mechanism} names the control that would sharpen the
interventions: vary which sibling the anchor points to while holding the
neighbour lists fixed. Running it needs new inference, but its premise is
checkable on the frozen release, and checking it fixes what the sibling
families can be said to measure, whether or not the intervention is ever
run.

\textbf{What the evidence actually supplies.} In the GraphSpace view of
every S4a, S4b and S4c case, the rendered evidence contains exactly one
typed neighbour list, \texttt{sibling\_neighbours\_typed}, belonging to
the one sibling that \texttt{orbit\_link} names. In every S2 case it
contains exactly one entry under
\texttt{target\_neighbours\_typed\_by\_site\_ref}, belonging to the one
$k$-th neighbour the question asks about. Across all $54{,}156$ cases
in these four families there is never a second candidate list. The
pointer therefore has no competing referent to discriminate, and
following it is not required to find the right list: there is
only one.

\textbf{A solver that never reads the pointer.} We wrote a
deterministic program with no learned parameters that ignores
\texttt{orbit\_link}, \texttt{sibling\_site\_ref}, the anchor
identifier and the question's $k$, reads only the supplied neighbour
list, groups its entries into shells by the integer part of the printed distance,
and returns the first-shell size (S4a, S2), the second-shell size (S4b), or
the modal element of the first shell (S4c). That grouping is not the release
builder's \texttt{distance\_bin}, which labels bands at $2$, $3$, $4$ and
$6$~\AA{}; the rule needs nothing but the printed numbers, which is what
makes it runnable by a reader who never sees the builder.
Table~\ref{tab:pointer-blind} reports what it scores.

\begin{table}[H]
\caption{\textbf{A pointer-blind list rule vs.\ majority baseline and model}, four sibling/two-hop families.}
\label{tab:pointer-blind}
\centering
\small
{\setlength{\aboverulesep}{0pt}\setlength{\belowrulesep}{0pt}%
\begin{tabular}{llrrrr}
\toprule
\rowcolor{HeaderColor}
\textbf{Family} & \textbf{Answer} & \textbf{Items} & \textbf{Majority} & \textbf{List rule} & \textbf{Model} \\
\midrule
S2  & CN of the $k$-th neighbour  & 23{,}993 & 27.2 & \textbf{94.9} & --- \\
S4a & CN via the sibling          & 10{,}197 & 31.6 & \textbf{88.1} & 31.8 \\
S4b & Second shell via sibling    & 9{,}769  & 22.4 & \textbf{99.6} & 18.6 \\
S4c & Dominant element via sibling& 10{,}197 & 50.5 & \textbf{93.2} & 68.4 \\
\bottomrule
\end{tabular}}
\end{table}
\smallskip
All figures are percentages. Majority is the modal-answer rate from the
release manifest. The model column is the frozen baseline checkpoint on
the GraphSpace view of the $1{,}900$-case subsample, scoring every
non-parsing output as an error; S2 is absent from that subsample.

\textbf{What follows.} The list rule solves these families. It beats the
model on all three families where both are measured; the model sits at the
majority baseline on S4a ($31.8$ against $31.6$) and below it on S4b ($18.6$
against $22.4$), though the majority column is computed over the full family
and the model column over the $1{,}900$-case subsample, so the two are not
paired. Whatever the
model is doing on the sibling families, the tasks themselves do not
require the anchor-to-sibling-to-neighbours composition their
construction was meant to force, because the composition's only
function is to select a list and no selection is ever needed.

This has a direct consequence for the interventions in
Table~\ref{tab:A1}. Deleting the typed neighbour shell removes the
single input that both a relational reader and the list rule depend on,
so the $-45.6$~pp crash is predicted by both accounts and separates
neither. Injecting that shell into a plain graph supplies it, so the
$+27.7$~pp recovery is likewise consistent with both. The small
$-5.0$~pp cost of removing the pointer fields, which we previously
found hard to interpret, is what one should expect when the pointer is
not needed to locate the list. We therefore read none of these
measurements as evidence of multi-hop inference, and
Section~\ref{sec:mechanism} states the narrower claim they support.

\textbf{Why the anti-shortcut protocol did not catch this.} The
release's anti-readback contract requires that the gold token never
appear adjacent to its answer key and that all symmetry, Wyckoff and
crystal-system fields be withheld, and the S4 families satisfy it: the
answer is computed from the evidence rather than printed in it. That
test rules out copying a visible answer. It does not rule out a short
deterministic computation over the visible evidence that reaches the
same answer by a different route than the intended one. A family is shortcut free only if no such program exists, and establishing that
requires running candidate programs, not inspecting the rendered text.
We treat the audit in this appendix as the missing gate, and a family
should be admitted to a reasoning claim only after a pointer-blind
program has been shown to fail on it. Building sibling families with
several candidate lists carrying different answers, so the pointer
is the only way to choose, is the concrete repair. We carried it out;
Appendix~\ref{app:pointer-forced} reports the rebuilt family and the
frozen checkpoint's result.

\subsection{Pointer-forced composition test}
\label{app:pointer-forced}

The audit above shows the S4 families can be solved without following
\texttt{orbit\_link}. It does not show what the model does when
following it is unavoidable. We rebuilt the sibling family so that it
is, and ran the frozen checkpoint on it.

\textbf{Construction.} Each case presents $k=4$ candidate sites with
their typed neighbour shells, whose first-shell dominant elements are
pairwise \emph{distinct}, so no two candidates yield the same answer.
Only \texttt{orbit\_link} identifies which candidate is the anchor's
symmetry-equivalent sibling. Every candidate carries the same centre
element, the anchor is excluded from all rendered shells, and site
identifiers are re-drawn per case from a shuffled range so position and
index carry no information. Three families are emitted: F1, the
$k=4$ task, in a \emph{pointer shown} and a \emph{pointer deleted} arm
over identical evidence; F2, twin pairs whose candidate evidence is byte identical and which differ
only in which candidate \texttt{orbit\_link} names, with the gold following
the pointer. F2 intervenes on the rendered reference while the candidate
lists stay fixed, so it tests whether the model follows the reference it is
given; we do not claim each redirected twin is itself a realisable crystal.
The third is F3, a single-candidate control whose evidence is byte-shape
identical to the released S4c family, which separates inability to do the
underlying computation from inability to resolve the pointer.

\textbf{Ceiling.} Eleven pointer-blind routes are measured on the released
pointer-forced pack itself, not argued: answer the first, last, longest,
shortest, lowest-index or highest-index candidate; the pooled element
mode; the element present in the most candidate lists; the nearest
neighbour's element; the printed centre element; and the single best
constant answer. Cases are admitted by a sieve that holds every route,
and the gold marginal, near $1/k$. On the released set ($n=882$) the
rules span $23.2$--$27.0\%$. We report the highest \emph{measured}
rule, $27.0\%$, rather than the nominal $25\%$, because balancing cannot
drive every route to exactly $1/k$ and the looser bar would overstate
the result. Four routes were missed at construction time, three of them found only after
the earlier ones prompted a systematic sweep, and three cleared $30\%$
before balancing. It is the maximum over the rules we thought to run, measured on the same
released set the sieve rebalanced, and it is not a proven upper bound: rules
failing does not establish that the next one succeeds, and three of them
were added only after an earlier version of this construction leaked, so the
set served as its own development set for those three. We therefore treat it as a
falsifiable threshold rather than a guarantee, and release the audit
script so it is recomputable on the exact data reported.

\begin{table}[h]
\centering
\caption{Pointer-forced families, frozen checkpoint, GraphSpace view.
Every case counts, and a non-parsing generation counts as wrong.
Intervals are material-clustered bootstraps over 2,000 resamples. The last column compares the lower bound against $27.0\%$, the highest of
the eleven pointer-blind rules measured in
Appendix~\ref{app:pointer-blind}.}
\label{tab:pointer-forced}
\begin{tabular}{llrrrc}
\toprule
\rowcolor{HeaderColor}
\textbf{Family} & \textbf{Arm} & \textbf{$n$} & \textbf{Accuracy} & \textbf{95\% CI} & \textbf{$>$ max} \\
\midrule
F1 multi-candidate ($k{=}4$) & pointer shown & 882 & 32.3\% & $[29.2,\,35.3]$ & yes \\
F1 multi-candidate ($k{=}4$) & pointer deleted & 882 & 26.8\% & $[23.9,\,29.5]$ & no \\
F2 pointer swap & pointer shown & 1{,}764 & 25.9\% & $[24.2,\,27.6]$ & no \\
F3 single-candidate control & pointer shown & 882 & 60.2\% & $[55.7,\,64.7]$ & --- \\
\bottomrule
\end{tabular}
\end{table}

\textbf{Result.} In Table~\ref{tab:pointer-forced} the checkpoint reaches
$60.2\%$ on F3, so it can
perform the underlying computation when the target is unambiguous. On
F1 it reaches $32.3\%$, above the $27.0\%$ measured maximum, and deleting the
pointer from identical evidence costs $5.56$~pp (paired,
$[+3.97,\,+7.03]$, $n=882$). The pointer is therefore not ignored. The margin is small against what is achievable: $32.3\%$ with four
candidates against $60.2\%$ when the target is unambiguous, both over the
same $n=882$. We quote those two measured rates rather than converting the gap into a
fraction of cases resolved, which would rest on a mixture model rather than
on a measurement.

F2 settles the mechanism. The model transcribes the pointer correctly:
it emits an explicit \texttt{orbit\_link} claim on $1{,}707$ of the
$1{,}764$ F2 cases ($97\%$; the F1 rate is $839$ of $882$, $95\%$), and
every one of those claims names the site the pointer actually names.
The reading below therefore covers the cases where the model's own
resolution is legible, which is almost all of them. Yet when only the
pointer moves and the candidate evidence is byte identical, $95.6\%$ of
twin pairs receive the \emph{same} answer, and both twins are correct
in $0.9\%$ of pairs. Which candidate the model actually computes over is spread across the four
render slots, at $231/229/178/230$ of the $868$ pairs whose resolution is
legible, the other $14$ of $882$ having no readable resolution. The pointer is copied into the output without
being used to select the evidence the answer is computed from.

\textbf{What this licenses.} The composition claim that
Section~\ref{sec:mechanism} withdrew is not recovered: on a family
where the pointer is load bearing, the frozen checkpoint performs near
the measured pointer-blind maximum. This does not bear on the
representation result of Section~\ref{sec:experiments}, which contrasts
GraphSpace against plain graph; the pointer-blind rule reads
GraphSpace's own typed fields and cannot be run on the plain-graph
side at all. What the test removes is a mechanism claim, not a
measured effect. Whether the GraphSpace representation makes this
composition \emph{learnable}, as opposed to already present, is a
separate question, which the next subsection answers.

\subsection{The same composition after matched supervision}
\label{app:pointer-forced-sft}

The frozen checkpoint copies the pointer without using it. That leaves
open whether the deficit is a missing capability or missing
supervision. We fine-tuned the base model on the pointer-forced
construction and re-ran the identical held-out test.

\textbf{Setup.} Base Qwen3.6-27B, not the main checkpoint, so a view
already trained on GraphSpace data cannot confound the outcome.
Answer-only supervision, no chain of thought: templated rationales have
repeatedly produced boilerplate emitted without doing the work, so the
only route above the measured maximum is the composition itself. Training set
$31{,}689$ rows from $1{,}510$ materials; test set $4{,}410$ rows from
$155$ materials, disjoint by material and verified row by row against
the training split. Every one of the eleven audited pointer-blind routes sits between $0.225$
and $0.253$ on the exact training bytes, so none of the routes we measured
is present in what is taught.

\begin{table}[h]
\centering
\caption{The same held-out pointer-forced test under two different
models, GraphSpace view. \textbf{Frozen} is the main union checkpoint
ckpt-14182. \textbf{Pointer-SFT} is the base Qwen3.6-27B trained
answer-only on pointer-forced cases from disjoint materials. These are
two models with different training histories, not two stages of one
run. Every case counts, and a non-parsing generation counts as wrong.
The highest of eleven measured pointer-blind rules reaches
$27.0\%$.}
\label{tab:pointer-forced-sft}
\begin{tabular}{llrrr}
\toprule
\rowcolor{HeaderColor}
\textbf{Family} & \textbf{Arm} & \textbf{$n$} & \textbf{Frozen} & \textbf{Pointer-SFT} \\
\midrule
F1 multi-candidate ($k{=}4$) & pointer shown   & 882   & 32.3\% & \textbf{99.8\%} \\
F1 multi-candidate ($k{=}4$) & pointer deleted & 882   & 26.8\% & 23.4\% \\
F2 pointer swap              & pointer shown   & 1{,}764 & 25.9\% & \textbf{99.9\%} \\
F3 single-candidate control  & pointer shown   & 882   & 60.2\% & 99.8\% \\
\bottomrule
\end{tabular}
\end{table}

\textbf{Result.} The composition is learnable
(Table~\ref{tab:pointer-forced-sft}). With the pointer shown
the model reaches $99.8\%$ against the $27.0\%$ measured maximum, and the paired
pointer effect rises from $+5.56$ to $+76.42$~pp
($[+73.70,\,+79.25]$, $n=882$). The twin diagnostic inverts: pairs
that differ only in which candidate the pointer names received the same
answer $95.6\%$ of the time when frozen and $0.11\%$ of the time after
training. The answer now follows the pointer.

Three checks establish that this is composition rather than a shortcut
outside the enumerated set. First, and most important, deleting the pointer
drops accuracy to $23.4\%$, \emph{below} the $27.0\%$ measured maximum: any rule that reached the answer without the pointer would work just as well
in this arm, so none of the eleven measured rules is what produces the
answer here. Second, the emitted answer equals the first-shell mode of the site the
pointer names on $99.8\%$ of F1 cases and $99.9\%$ of F2 cases. That
equality is the task's own correctness condition, so it shows which
candidate list the answer is computed over without identifying the
computation that runs on it. Third, the answer equals the mode of the \emph{first rendered} candidate on
$25.9\%$ and $25.1\%$ of cases, which is chance for $k{=}4$ and leaves no
room for a positional rule.

\textbf{What this licenses, and what it does not.} It establishes that
the composition is learnable in this representation and that the
trained model's answer is pointer dependent. It does not establish that
GraphSpace is \emph{necessary} for that: the matched plain-graph arm
was not run, because the plain-graph rendering of these families does
not carry the information the questions require
(Appendix~\ref{app:view-parity}). This is one seed and one arm; Appendix~\ref{app:consolidated} sets out the
several arms and the replication across seeds that would close it. Training and test share the task
construction and differ in materials, so this is generalisation across
materials, not across tasks. Training loss is not evidence here: it fell
to $1.5\times10^{-5}$ within the first epoch and to $5.7\times10^{-7}$
by the end, which memorisation of $31{,}689$ rows by a 27B model
explains on its own, so the result rests entirely on the held-out set and
not on how closely the training rows were fit.

\subsection{Information parity between the two evidence views}
\label{app:view-parity}

A view comparison measures representation only if both views carry what
the answer needs and differ in presentation. Applied to our own healthy
pool, three of seven families fail that test.

\textbf{What the plain view omitted.} For S4a, S4b and S4c the
plain-graph evidence carried no element labels, no distances on its
edges, and no equivalence link, on every one of the $30{,}163$ cases in
those families. Those questions instruct the model to reach a masked
anchor through a symmetry-equivalent sibling. Without the link the
sibling cannot be identified; without distances no shell can be
segmented; without element labels no element can be named. An oracle that solves each case directly from the rendered plain-graph
evidence scores $0.000$ on both S4a and S4c, failing on all $10{,}197$ cases
of each for want of a pointer. The rendering therefore lacks the fields the
intended solution path needs; it does not by itself rule out some other
route to the same answer.

\textbf{Consequence for the measured gap.} Table~\ref{tab:view-parity}
splits the headline delta by whether both views carry the required
inputs. It is the same artefact, checkpoint set, seeds and denominator
as Table~\ref{tab:healthy-deltas}: the GraphSpace-full training view,
pooled over seeds 42/43/44, and the two groups sum back to
$n=229{,}485$ and $+0.1933$. On the three families whose plain view
cannot answer, the gap is $+46.7$~pp. On the four where both views
carry the information, it is $+1.96$~pp. The matched size of the representation effect on this pool is therefore
about two points rather than nineteen, and it varies by family: S1, S3 and S4d favour GraphSpace with
intervals excluding zero, while S2 favours the plain rendering. The group rows are $n$-weighted means of the family rows, and S2 carries
most of the weight in the matched group.

Plain-graph accuracy of $0.009$ on S4a and $0.032$ on S4b, measured on the
frozen main checkpoint, sits far below those families' own majority
baselines of $0.316$ and $0.224$ (Table~\ref{tab:pointer-blind}), which a
rendering that merely made the task harder would not require. That is
consistent with a missing input, and the solvability oracle below tests it
directly.

\begin{table}[h]
\centering
\caption{\textbf{The headline view gap split by information parity.}
Same artefact as Table~\ref{tab:healthy-deltas}: the GraphSpace-full
training view over seeds 42, 43 and 44, paired on case id. The two
groups sum to the 229,485 items and the $+0.1933$ quoted in
Section~\ref{sec:experiments}. Family intervals are the pooled paired
bootstrap. Group rows are weighted by item count and carry no
re-bootstrapped interval.}
\label{tab:view-parity}
\begin{tabular}{llrrl}
\toprule
\rowcolor{HeaderColor}
\textbf{Group} & \textbf{Family} & \textbf{$n$} & \textbf{$\Delta$} & \textbf{95\% CI} \\
\midrule
Both views carry   & S1 orbit count          & 19{,}401 & $+0.0764$ & $[+0.0717,+0.0814]$ \\
the required       & S2 $k$-th nbr.\ CN      & 71{,}163 & $-0.0158$ & $[-0.0182,-0.0133]$ \\
inputs             & S3 argmax elem.\ coord  & 20{,}862 & $+0.0842$ & $[+0.0791,+0.0895]$ \\
                   & S4d multiattr match     & 28{,}971 & $+0.0218$ & $[+0.0201,+0.0237]$ \\
\rowcolor{HeaderColor!30}
\textbf{pooled}    &                         & \textbf{140{,}397} & $\mathbf{+0.0196}$ & --- \\
\midrule
Plain view cannot  & S4a anchor CN           & 30{,}107 & $+0.5915$ & $[+0.5853,+0.5974]$ \\
answer: no pointer, & S4b second shell       & 28{,}873 & $+0.5416$ & $[+0.5360,+0.5474]$ \\
no distances, no   & S4c dominant element    & 30{,}108 & $+0.2711$ & $[+0.2637,+0.2780]$ \\
element labels     &                         &          &           &  \\
\rowcolor{HeaderColor!30}
\textbf{pooled}    &                         & \textbf{89{,}088}  & $\mathbf{+0.4670}$ & --- \\
\bottomrule
\end{tabular}
\end{table}

A second asymmetry compounds it: the old plain prompts were an order of
magnitude longer than their GraphSpace counterparts and were truncated
on $25.9\%$ of rows against $6.6\%$, which depresses them for a reason
also unrelated to representation.

\textbf{Why our own protocol did not catch it.} The anti-readback
contract requires that the gold token never be printed next to its
answer key, and the S4 families satisfy it. That contract tests whether
an answer is \emph{visible}. It never tests whether an answer is
\emph{reachable}, and those are different properties. The evidence was in our own build
artefacts, in two separate places. The release manifest records, per
family, that a ``GraphSpace advantage [is] expected (only gs has orbit
link + typed sibling neighbours)'', which states the information
asymmetry outright. Independently, the scoring script enforces a
\emph{truncation}-parity guard, failing when the two views' non-stop
rates differ by $0.02$ or more; on this comparison they differ by
$0.193$ ($0.066$ against $0.259$), so the guard fired, wrote
\texttt{parity\_ok: false} into the verdict and returned a non-zero
exit status. The number was read anyway. Neither signal was about the
other: one names the missing inputs, the other flags a length
artefact. Both were computed, recorded and then not propagated
into the claim. We now gate on two checks that
must both pass before any accuracy is read from a view comparison: a
parity check that each required input is present in both views, and a
solvability oracle that derives the gold from each view's own rendered
evidence. The first alone is insufficient: an intermediate repair
passed it while S4a remained unsolvable, because presence of an input is
not the same as recoverability of the answer. On the rebuilt release both
checks pass and the oracle solves the two families it covers at identical
rates from either view ($0.8813$ on S4a and $0.9322$ on S4c, $n=10{,}197$
each, with every missing-input counter at zero in both), so neither view
is advantaged by an information gap; the residual is the oracle's
shell-segmentation rule, which bins by $\lfloor\text{distance}\rfloor$,
not the release's rule.

\textbf{What the rebuilt evaluation shows, and what it cannot settle.}
Evaluating the rebuilt release on both views over $7{,}000$ paired cases
confirms the repair and leaves the reported figures unchanged. Its GraphSpace prompts are byte identical to the old release on six of the
seven families, S2 excepted, so no movement on that side comes from the
prompts; the serving configuration did change, and the control below bounds
what that is worth. Truncation parity is restored: the pooled non-stop gap
falls from $0.193$ to $0.017$, and the plain view's truncation on S4a,
S4b and S4c falls from $31.7\%$, $48.4\%$ and $47.4\%$ to $2.6\%$,
$0.8\%$ and $0.4\%$, so the old plain view was not merely uninformative
on those families but degenerating. The three-family advantage falls
from $+14.2$ to $+6.8$~pp. Two findings stop us reporting the rebuilt
numbers in place of the published ones. The run was served at
tensor-parallel $2$ rather than $8$, and a control puts that change at
$-19$~pp on S4a and $-16$~pp on S4c against a $\pm 0.1$~pp determinism
floor, which is larger than the effect in question. And the rebuild
induced a scoring artefact: the plain view now returns a correct answer
beside a malformed \texttt{evidence\_refs} list, the field the repair
added, on $687$ of $1{,}000$ S4c cases, $540$ of them correct. Reading
the answer field alone over the same denominator, still counting a
missing answer wrong, moves S4c from $+0.263$ to $-0.255$. A delta whose
sign turns on a parser convention is not a representation effect. What survives every combination of release, weighting and parser is the
matched-family result: eight such estimates fall between $-3.9$ and
$-0.7$~pp, agreeing with the $+1.96$~pp of the published run in placing the
matched-family effect within a few points of zero rather than near nineteen.

\subsection{Paired subsets in the thinking-supervision table}
\label{app:paired-subsets}

Not every generation parses into valid JSON with an answer field. The
frozen baseline emits a parseable answer on $1{,}835$, $1{,}900$,
$1{,}680$, $1{,}815$ and $1{,}574$ of the $1{,}900$ cases for the five
rows of Table~\ref{tab:A2v6} in table order, and the
thinking-supervised model on $1{,}542$, $1{,}547$, $1{,}526$,
$1{,}403$ and $1{,}326$. The supervised model loses most under the
sibling-reference mask ($1{,}403$ of $1{,}900$, $73.8\%$), because it
emits a thinking block before the answer and more often reaches the
$16{,}384$-token cap; the frozen baseline loses most under the
typed-shell mask ($1{,}574$, $82.8\%$).

An earlier version of this table used the scorer's predicted-only
mode, which drops a case unless both arms parsed and so gives each row
its own denominator ($1{,}528$, $1{,}547$, $1{,}392$, $1{,}377$ and
$1{,}096$). That rule conditions on an outcome the intervention itself
affects, so we no longer report it as the main result.
Table~\ref{tab:A2v6} instead scores every non-parsing output as an
error over the common denominator of $1{,}900$, with
material-clustered paired bootstrap intervals ($B = 2{,}000$, seed
$20260820$, $1{,}565$ clusters). All five contrasts keep their sign
and every interval still excludes zero, but the three positive rows
shrink by roughly half, from $+7.13$, $+7.89$ and $+7.76$~pp to
$+3.16$, $+3.89$ and $+5.84$~pp, while the two negative rows move slightly, one away from zero and one toward
it, from $-10.24$ and $-8.49$ to $-11.16$ and $-7.89$~pp. Output failure is therefore a real component of the effect
size on the positive rows. The conclusion that thinking supervision
helps on grounded input and hurts under masking survives the stricter
accounting; its magnitude on grounded input does not.

\section{Consolidated claims and outlook}
\label{app:consolidated}

This appendix states what the three analysis lines jointly establish, what
they leave open, and what would settle each open item. It repeats no
measurement: every number below is reported with its protocol in the section
named beside it.

\subsection{What the three lines jointly establish}

\textbf{The dependence is on the typed list, not on the traversal.} The
reveal attack and the pointer-blind audit meet on the same object from
opposite directions. Causally, deleting the typed-neighbour shell drops S4c
from $70.7\%$ to $25.1\%$, a fall of $45.6$~pp with interval
$[-51.7,-39.6]$, and injecting the same typed edges into the plain-graph
view raises paired S4c from $46.6\%$ to $74.3\%$, a gain of $27.7$~pp with
interval $[+21.8,+33.9]$. A single leg would be consistent with a formatting effect; for two legs
moving in opposite directions on the same field, a formatting account has to
act on that field in both directions, which is a narrower proposal than a
general sensitivity to layout.
Correlationally the same field is implicated: grounded cases that cite both
an \texttt{orbit\_link} and a \texttt{sibling} literal answer S4c at
$72.1\%$ against $20.0\%$ for cases citing neither, while in the plain view
the \texttt{orbit\_link} citation rate is exactly $0.000$ because the field
does not exist there. The pointer-blind audit then shows those lists are not
only necessary but sufficient: a parameter-free rule that follows no pointer
reaches the answer on four of the seven hardened families. Necessity and
sufficiency for the same field is why we read the dependence as resting on the list itself rather than on
traversal between its entries, which is the narrower of the two readings
(\S\ref{sec:mechanism}, Appendix~\ref{app:pointer-blind}).

\textbf{The composition is learnable from this representation.} The
pointer-forced rebuild separates citing a relation from using it. Frozen,
the checkpoint reaches $60.2\%$ when the list is unambiguous, gains only
$5.56$~pp from being shown the pointer, and returns the same answer for
$95.6\%$ of the pairs in which nothing moves but the pointer: it transcribes
the pointer without conditioning on it. Matched supervision behaves
differently at every reading: $99.8\%$ with the pointer, twin agreement
$0.11\%$, and $23.4\%$ once the pointer is deleted, which is \emph{below}
the $27.0\%$ measured shortcut maximum. That last number is the decisive
one, because a rule reaching the answer without following the pointer would
have scored at or above the maximum. What this establishes is that the
composition is learnable from this representation. It does not establish
that the representation is required for it, which is a separate arm
(\S\ref{app:necessity} below).

\textbf{The formal properties are audit results, not theorems.}
Canonicalisation is idempotent and the round trip is lossless on every
audited row, the hierarchical refinement raises zero warnings and no
collision, the v2 edge-case wrapper passes $47/47$, and training on the
canonical form rather than the expanded one is worth
$\Delta_{\mathrm{IO}} = +11.77$~pp with interval $[+11.46,+12.06]$. The
weakest cell is elemental substitution at $82.5\%$, where seven anomalies
track species-dependent bond lengths. These are properties of the audited
releases under the stated build, not a uniqueness result over crystals in
general, and we state them that way throughout
(\S\ref{sec:formalization}).

\textbf{The two lines meet at CE-2.} The field that the injectivity argument
depends on is exactly the field whose removal produces the $-45.6$~pp mask
and whose injection produces the $+27.7$~pp gain. The formal content of the
representation and the behavioural dependence of the model are, at that
point, about the same object. This is the strongest internal consistency
check in the paper, and it is the one we would most want an independent
group to reproduce.

\subsection{What is not established, and what would settle it}
\label{app:necessity}

\textbf{The reader circuit stays a candidate.} Mechanistic tracing names
candidate reader heads, but the pooled reader-versus-random ablation
interval includes zero at $+1.82$~pp, $[-0.20,+3.98]$ over three seeds.
Naming a head is not localising a computation. Settling this needs an excess
effect over random-head ablation at an effect size fixed in advance, held
across seeds, with the head set chosen on data disjoint from the data it is
scored on.

\textbf{Necessity of GraphSpace is untested.} Showing that GraphSpace is
\emph{necessary} for this composition, rather than sufficient for it,
requires an information-matched plain-graph training arm on the rebuilt
view. That arm's evaluation side is built and validated: the solvability oracle
equalises the two views and pooled truncation parity moves from $0.193$ to
$0.017$. Appendix~\ref{app:view-parity} reports that rebuilt run separately from the
published one, and necessity is withheld rather than argued until a training arm is matched to
that evaluation side.

\textbf{The global rule refuses, reproducibly.} The broader decision rule is
not admitted in any of the three seeds, because EFS-path regression against
the best competing view breaches the $-2$~pp tolerance every time. This is a stable refusal rather than noise, and the two routes forward are
to improve that family or to renegotiate the tolerance before the fact,
never after seeing the seeds (\S\ref{sec:certification-analysis}).

\textbf{Scale is not the remedy.} Across four measured training sizes on one
18,000-query protocol the representation effect changes sign and the
supervision effect peaks at an intermediate size rather than the largest.
More data, by itself, does not stabilise these effects.

\textbf{Answer visibility bounds what any view score means.} Removing the
symmetry/local block costs $28.8$ points and drops GraphSpace below the
raw-CIF view, and the layer-48 probe follows the same pattern, with an
unweighted mean of $41.2\%$ when the printed field is withheld against
$65.1\%$ when it is present. No score from an answer-printing view is a
capability claim, which is why those cases are diagnostics here and are
excluded from the hardened metric.

\subsection{Reading for materials practice}

The defensible reading is narrow and still useful. Explicit typed relations
can make structural evidence available to a model that would otherwise not
recover it from a flat serialisation, and that evidence changes predictions
on the tested neighbour tasks. On families of that shape, structured inputs
are worth preferring. The reading does not extend to materials reasoning in
general: the hardened families are synthetic structure-reasoning tasks with
the answer token withheld by construction, the evidence comes from a single
27B checkpoint, and on the external transfer panel GraphSpace sits at
ceiling only where the answer is printed. A practitioner should read this
paper as a statement about evidence format on a characterised class of
questions, and should re-run the anti-shortcut battery before assuming the
class covers their own tasks.

\subsection{Future work}

The pointer swap this paper identified as missing has been run
(\S\ref{sec:mechanism}, Appendix~\ref{app:pointer-forced}); what it leaves
open is the matched plain-graph arm, which the old plain rendering could not
support. Closing it needs the information-matched training arm described
above, plus replication across seeds and arms, and a re-run at the published
serving configuration, so that the serving change itself cannot confound the comparison, which
Appendix~\ref{app:view-parity} documents together with the parity repair it
required.

Three extensions follow from results already in hand. A full-SFT
thinking-augmentation run (\hyperref[sec:A2]{the thinking-supervision
analysis}) would test whether an explicit thinking head over the same typed
relations tightens the reveal-attack mask-replay signature; the current
thinking arm gives a paired IO gain of $+3.16$~pp on the grounded view, CI
$[+0.79,+5.46]$, which is positive but too small to carry a mechanism claim.
The hierarchical canonical form (\hyperref[sec:B6]{the hierarchical-form
audit}) is a second starting point for large-supercell and MOF-scale
materials, since (canonical.\-flat, canonical.\-hierarchical) already forms
a proper refinement and the two can be trained jointly. The v2 schema
(supercells, defects, disorder, solid solutions) extends the same controls
to defect-driven prediction once symmetry-clean crystals are handled; the edge-case wrapper passing $47/47$ is what makes that extension testable
on the current releases rather than a plan for later.

Two methodological items are worth carrying to other benchmarks
independently of GraphSpace. The first is the solvability oracle: gating a
view comparison on whether both views can answer at all turned a
$+19.3$~pp headline into $+1.96$~pp, and no representation result is
interpretable without that gate. The second is running the shortcut battery
against one's own hardened set rather than only against the baseline, which
is what surfaced the pointer-blind rule on four of our seven families.

\section{Training hyperparameters}
\label{app:training-hparams}

Every SFT arm in the P8 sweep starts from the same public base
checkpoint, \texttt{Qwen/Qwen3.6-27B}, loaded in \texttt{bfloat16}
from a local mirror and fine-tuned with all parameters trainable; no
adapter, quantization, or parameter freezing is used anywhere in the
main sweep. The model is dense, not a mixture of experts, and the
analysis in Section~\ref{sec:mechanism} traces its $64$ decoder
layers. The mechanism experiments of Appendix~\ref{app:A2} start
instead from the union-SFT checkpoint
\texttt{graphspace\_main\_v1\_union\_...\,/checkpoint-14182}, which is
itself a full-parameter fine-tune of the same base.

Every SFT arm in the P8 sweep shares the schedule described here.
Optimization uses Adam with peak learning rate $5\text{e-}6$, linear
decay to $5\text{e-}7$, warmup over the first $3\%$ of steps, and
weight decay $0.1$. The global batch size is $64$, packed sequence
length is $16{,}384$, and tensor-parallel degree is $8$. Each arm
trains for $3$ epochs on $65{,}536$ rows and validates on $4{,}096$
matched rows. A single material-identity hash assigns train,
validation, and test partitions across all views, and \texttt{eval\_id}
links the rendered questions across views for paired comparison.
Twenty-four arms cover the eight views crossed with seeds $42$, $43$,
and $44$. Training YAMLs and per-view manifest checksums are archived
in the artefact repository (Appendix~\ref{app:uuid-ledger}).

\section{Reproducibility}
\label{app:uuid-ledger}

Every mechanism and representation-level experiment reported in the
paper is scored by one frozen scorer and pinned by input-pack SHA-256
checksums. This appendix names the entry points needed to recompute
each headline number; the repository accompanying the paper mirrors
these paths unchanged, so a path given here resolves in both places.

\textbf{Pool admission.} The four public sources behind the vetted pool are
the Materials Project \citep{jain2013materialsproject}, OMat24, OpenDAC and
QMOF, with no source exceeding $31\%$ of the pool. Admission verifies
structure legality, deduplicates by the composition, space-group and
Wyckoff-site identity of \S\ref{sec:data}, and requires the gold labels
evaluation needs.

\textbf{Reported-value check.} A final consistency check recomputes every
interval independently, compares it digit for digit with the reported value
and halts on a mismatch, with an adversarial unit test injecting tampered
scores to verify that the check catches them.

\textbf{Frozen releases.} The $1{,}024$-question benchmark and the
readback-hardened pool are separate frozen releases.
The pool is
\texttt{crystal\_reasoning\_eval\_hardened\_v4\_20260814} under
\texttt{data/releases/graphspace\_nature/},
whose \texttt{manifest.json} carries family counts, majority
baselines, per-view row counts, and the two checksums
\texttt{6de3f370e7fa\ldots} (pre-canonical) and
\texttt{96289ffb1d10\ldots} (with canonical spec). The manifest also
records \texttt{training\_allowed: false}, so no evaluation release was
admitted into any training pack.

\textbf{Canonical specification.} The canonical form audited in
Section~\ref{sec:formalization} is
\texttt{canonical\_version} v1: three-key site order
\texttt{wyckoff\_letter} then \texttt{periodic\_number} then
\texttt{frac\_coord\_lex}, a $6.0$~\AA{} neighbour cutoff,
\texttt{symprec} set explicitly to $10^{-3}$ rather than \texttt{spglib}'s
own default of $10^{-5}$, and
multi-edge compression by summed multiplicity. The reference
implementation is
\texttt{src/materials\_llm/graphspace/canonical.py} and the written
specification is \texttt{docs/graphspace\_canonical\_spec\_v1.md}.

\textbf{Statistics.} All paired intervals in the paper are produced by
\texttt{src/materials\_llm/graphspace/p8\_paired\_statistics.py},
which also defines the preregistered primary metric. Bootstrap seeds
are fixed per experiment and recorded in the verdict log: $20260715$ for the benchmark decision rule, $20260813$ for the
healthy-pool matrix, $20260820$ for the mechanism and
thinking-supervision contrasts, $20260830$ for the mask replay.

\textbf{Verdict logs.} Each experiment writes one verdict log holding
its preregistered rule, its inputs, and its outcome including
negatives. The healthy-pool matrix is
\texttt{p8\_final\_matrix\_verdict\_20260817.md} under
\texttt{docs/logs/graphspace\_nature/},
the thinking-supervision contrast is
\texttt{a2\_v6\_supervision\_verdict\_20260821.md}, the circuit trace
is \texttt{a3\_3seed\_pooled\_verdict\_20260820.md}, the canonical
ablation is \texttt{b5\_canonical\_ablation\_verdict\_20260819.md},
and the reveal attack is
\texttt{a1\_reveal\_attack\_verdict\_20260821.md}. Each log records
its decision rule before its results, so a rule that was not met can
be checked against the rule as written; the EFS-path tolerance breach
and the inconclusive three-seed circuit contrast are both reported
that way.

\textbf{Admission and certification.} The five certification criteria
of Section~\ref{sec:method} are evaluated by
\texttt{src/materials\_llm/graphspace/p8\_decision\_contract.py},
which refuses any statistics file whose
\texttt{primary\_metric} is not the preregistered one. Admission of a
release into training is a separate gate: a release carries
\texttt{training\_allowed} in its manifest, and every evaluation
release used in this paper carries \texttt{false}.

\textbf{Training arms.} Write
$R = \texttt{p8\_8view\_16k\_ep3\_seed}\langle s \rangle\texttt{\_fixed\_20260801}$
for $s \in \{42, 43, 44\}$ and $v$ for one of the eight views. The
$24$ training arms live under \texttt{train/output/graphspace\_}$R/v$,
each with an \texttt{args.json} recording the base model, seed, epoch
count, and sequence length. The matching inference runs live
under \texttt{eval\_runs/graphspace\_nature/}$R/v$, in four shards
\texttt{shards/shard\_0} through \texttt{shard\_3}, each writing an
\texttt{eval\_records.jsonl} whose rows carry
\texttt{answer\_field\_accuracy}, \texttt{hard\_slice},
\texttt{task\_family} and \texttt{split\_group\_id}. Every number in
Tables~\ref{tab:overall}, \ref{tab:family}, \ref{tab:metric-map}
and~\ref{tab:slice-map} aggregates those four files for one view.

\end{document}